\RequirePackage{fix-cm}
\documentclass[twocolumn]{svjour3}          

\smartqed           
\journalname{International Journal of Computer Vision}

\usepackage{graphicx}
\usepackage{mathptmx}      
\usepackage{latexsym}
\usepackage{times}
\usepackage{epsfig}
\usepackage{graphicx}
\usepackage{amsmath}
\usepackage{amssymb}
\usepackage{indentfirst}

\usepackage{bbold}
\usepackage{dsfont}

\usepackage{enumerate}
\usepackage{enumitem}
\usepackage{xspace}
\usepackage{xcolor}

\usepackage[toc,page]{appendix}
\usepackage[font=small,labelfont=bf]{caption}

\usepackage{xcolor}
\usepackage{amsfonts}

\usepackage{caption}
\usepackage{subcaption}
\usepackage{bm}
\usepackage{isomath}

\usepackage[numbers]{natbib}

\usepackage[pagebackref=true,breaklinks=true,letterpaper=true,colorlinks,bookmarks=false]{hyperref}
\usepackage{footmisc}
\usepackage{stmaryrd}

\usepackage{fixltx2e}
\usepackage{dblfloatfix}
\usepackage{pbox}
\usepackage{capt-of}

\usepackage[normalem]{ulem}
\usepackage{multirow}

\usepackage{colortbl}

\usepackage{array}

\usepackage{colortbl}
\usepackage{tabularx}
\usepackage{arydshln}
\usepackage{xspace}

\newcommand{\comment}[1]{}
\newcommand{\hg}[1]{{#1}}

\makeatletter
\@namedef{ver@everyshi.sty}{}
\makeatother

\definecolor{pink}{HTML}{db5a6b}
\definecolor{lblue}{HTML}{2e4e7e}
\definecolor{tiffany}{HTML}{1bd1a5}

\makeatletter
\DeclareRobustCommand\onedot{\futurelet\@let@token\bmv@onedotaux}
\def\bmv@onedotaux{\ifx\@let@token.\else.\null\fi\xspace}
\def\eg{\emph{e.g}\onedot} 
\def\ie{\emph{i.e}\onedot} 
 
\def\etc{\emph{etc}\onedot} \def\vs{\emph{vs}\onedot}
\def\wrt{w.r.t\onedot} 
\def\etal{\emph{et al}\onedot}
\makeatother

\begin{document}

\title{Event-guided Multi-patch Network with Self-supervision for Non-uniform Motion Deblurring}


\author{Hongguang Zhang         \and
        Limeng Zhang         \and
        Yuchao Dai         \and
        Hongdong Li           \and
        Piotr Koniusz 
}

\institute{H. Zhang (the corresponding author) is an Assistant Professor at the Systems Engineering Institute, AMS, Beijing, 100141, China.\at
           \email{zhang.hongguang@outlook.com}          
           \and
           L. Zhang is a PhD student at Shanghai Jiao Tong University.
           \and
           Y. Dai is a Professor at the Northwestern Polytechnical University.
           \and
           H. Li is a Professor at the Australian National University.
           \and
           P. Koniusz is a Senior Research Scientist in Data61/CSIRO and a Senior Honorary Lecturer at the Australian National University. \at
           \email{piotr.koniusz@data61.csiro.au}
}

\date{Received: 04.09.2021 / Accepted: 20.10.2022}

\maketitle

\begin{abstract}
Contemporary deep learning multi-scale deblurring models suffer from many issues: 1) They perform poorly on non-uniformly blurred images/videos; 2) Simply increasing the model depth with finer-scale levels cannot improve deblurring; 3) Individual RGB frames contain a limited motion information for deblurring; 4) Previous models have a limited robustness to spatial transformations and noise. Below, we extend our preliminary paper \cite{Zhang_2019_CVPR} by several mechanisms to address the above issues: I) We present a novel self-supervised event-guided deep hierarchical {Multi-patch Network (MPN)} to deal with blurry images and videos via fine-to-coarse hierarchical localized representations; II) We propose a novel stacked pipeline, StackMPN, to improve the deblurring performance under the increased network depth; III) We propose an event-guided architecture to exploit motion cues contained in videos to tackle complex blur in videos; IV) We propose a novel self-supervised step to expose the model to random transformations (rotations, scale changes), and make it robust to Gaussian noises. Our MPN achieves the state of the art on the GoPro and VideoDeblur datasets with a 40$\times$ faster runtime compared to current multi-scale methods. With 30ms to process an image at 1280 $\times$720 resolution, it is the first real-time deep motion deblurring model for 720p images at 30fps. For StackMPN, we obtain significant improvements over 1.2dB on the GoPro dataset by increasing the network depth. Utilizing the event information and self-supervision further boost results to 33.83dB.
\end{abstract}

\section{Introduction}
The objective of non-uniform blind image deblurring is to remove the undesired blur caused by the camera motion and the scene dynamics \cite{nah2017deep,tao2018scale,pan2017simultaneous}. 
Prior to the success of deep learning, conventional deblurring methods used to employ a variety of constraints or regularizations to approximate the motion blur filters, involving an expensive non-convex non-linear optimization, and overly restrictive assumption of spatially-uniform blur kernel, resulting in a poor deblurring of complex blur patterns.

\begin{figure}[t]
	\centering
	\includegraphics[height=5cm]{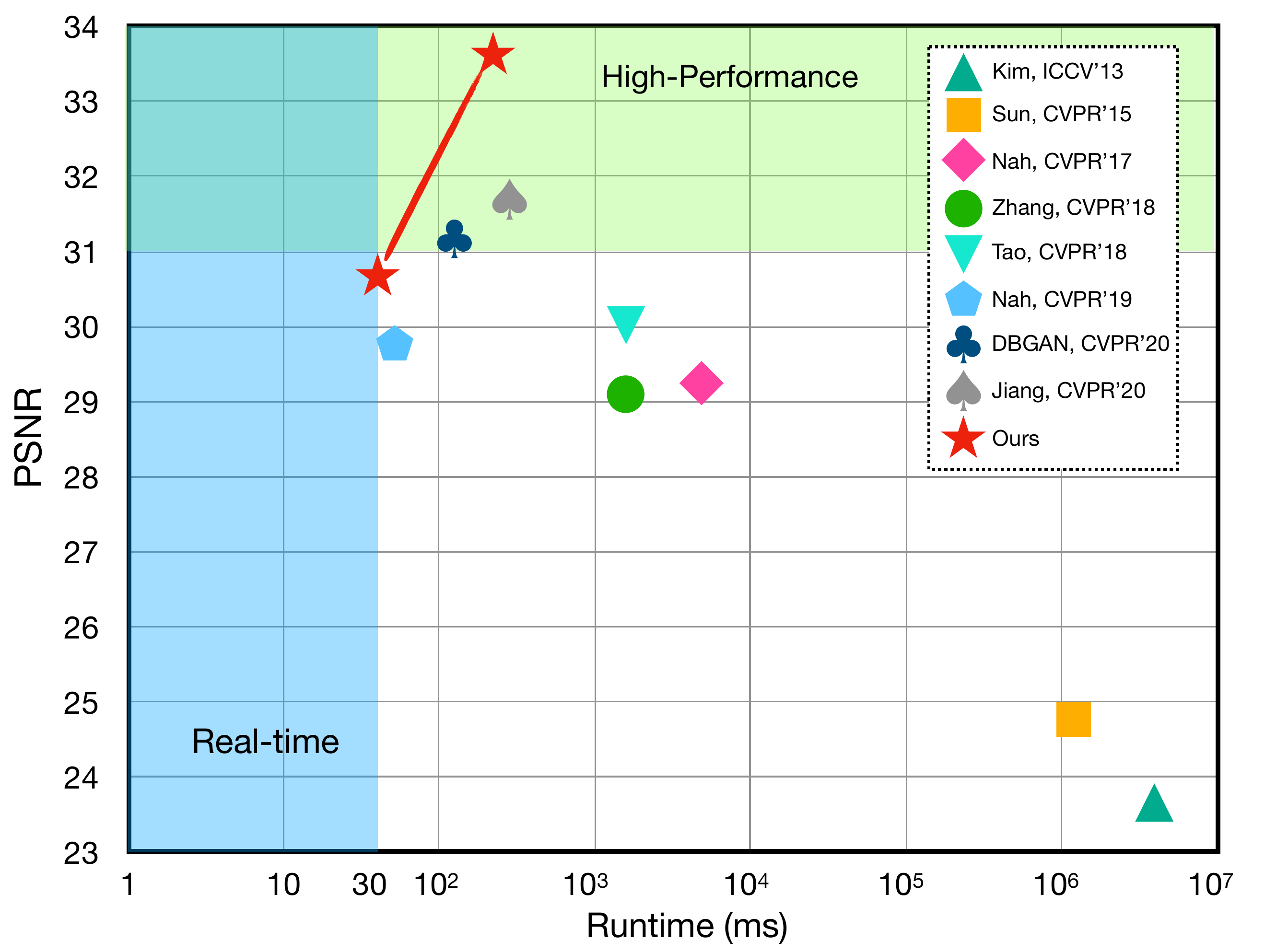}
        \caption{\small The PSNR vs. runtime of state-of-the-art deep image deblurring methods and our method on the GoPro dataset \cite{nah2017deep}. The blue region indicates real-time inference, whereas the green region represents high performance motion deblurring (over 30 dB). Our method achieves the best performance at 30 fps for $1280\!\times\!720$ images. The event-guided version and the stacked variants of our model further improve the performance at a cost of somewhat increased runtime. }
	\label{fig:PRPlot}
\end{figure}

\begin{figure*}[t]
	\includegraphics[width=\linewidth]{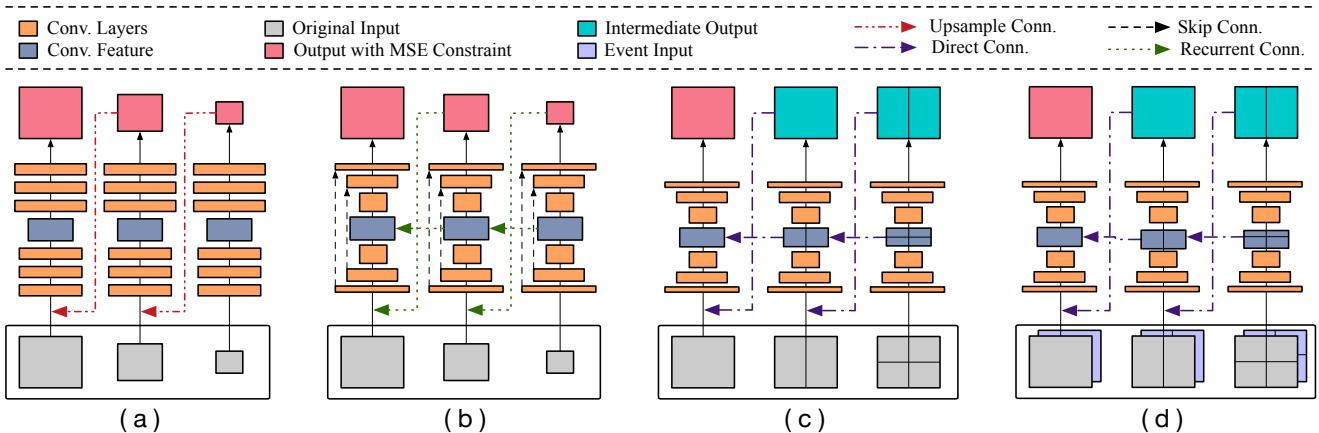}
        \caption{\small Comparison between different network architectures. From left to right: (a) multi-scale \cite{nah2017deep}, (b) scale-recurrent \cite{tao2018scale}, (c) our hierarchical multi-patch architecture. We do not employ any skip or recurrent connections which simplifies our model. (d) our event-guided multi-patch network architecture, in which the event representations are concatenated with original blurry frames as two-stream inputs. Best viewed in color.}
	\label{fig:Structure_Comp}
\end{figure*}

Deblurring methods based on deep Convolutional Neural Networks (CNNs) \cite{krizhevsky2012imagenet,simonyan2014very} learn the regression between a blurry input image and the corresponding sharp image in an end-to-end manner \cite{nah2017deep,tao2018scale}.
To exploit deblurring cues at varying processing levels, the ``coarse-to-fine'' scheme has been extended to deep CNN scenarios by a multi-scale network architecture \cite{nah2017deep} and a scale-recurrent architecture \cite{tao2018scale}. Under the ``coarse-to-fine'' scheme, a sharp image is gradually restored at different resolutions in a pyramid.
Nah \etal \cite{nah2017deep} demonstrated the ability of CNN models in removing motion blur from multi-scale blurry images, where a multi-scale loss function is devised to mimic conventional coarse-to-fine approaches. 
Following a similar pipeline, Tao \etal \cite{tao2018scale} shared network weights across scales to improve training and model stability, thus achieving highly effective deblurring compared with \cite{nah2017deep}. 
However, there still exist major challenges in current deep deblurring methods:

\begin{itemize}
    \item Under the coarse-to-fine multi-scale scheme, most networks use a large number of training parameters due to large filter sizes. Thus, the multi-scale and scale-recurrent methods suffer from an expensive runtime (see  Fig.~\ref{fig:PRPlot}) and struggle to improve the deblurring quality.
    \item Increasing the network depth for a low-resolution input in multi-scale approaches  does not seem to improve the deblurring performance \cite{nah2017deep}.
    \item The model is not capable of capturing motion information from RGB frames under complex blur, thus they cannot effectively address video deblurring.
    \item The learnt model has limited robustness to spatial transformations and random noises, which limits its usefulness in real-world applications. 
\end{itemize}

In this paper, we address the challenges with the multi-scale and  scale-recurrent architectures. We propose a novel architecture which exploits  deblurring cues at different scales via a \emph{hierarchical multi-patch} model.  Specifically, we propose a simple yet effective multi-level CNN model called deep Multi-Patch Network (MPN) which uses multi-patch hierarchy as input. In this way, the residual cues from deblurring local regions are passed via residual-like links to the next level of network which deals with coarser regions.  Feature aggregation over multiple patches has been used in image classification \cite{lazebnik2006beyond,he2014spatial,lu2015deep,koniusz2018deeper}. For example, \cite{lazebnik2006beyond} proposes Spatial Pyramid Matching (SPM) which divides images into coarse-to-fine grids in which histograms of features are computed. In \cite{koniusz2018deeper}, a second-order fine-grained image classification model uses feature embeddings of overlapping patches and  positional embeddings for aggregation. Sun \etal \cite{sun2015learning} learn a patch-wise motion blur kernel through a CNN for restoration via an expensive energy optimization.

The advantages of our network are threefold: 1) As the inputs at different levels have the same spatial resolution, we apply residual-like learning which requires smaller filter sizes and leads to a fast inference; 2) We use an SPM-like model  exposed to more training data at the finest level due to relatively more patches available for that level; 3) Our architecture encourages model to learn to deblurring from easier tasks (small patches) to harder tasks (large patches), a gradual learning process that encourages the consistency of deblurring over different locations and spatial sizes.

To overcome the limitation of {\em stacking depth} in  multi-scale and multi-patch models, simply increasing the model depth by introducing additional coarser or finer grids cannot improve the overall deblurring performance of known models. Thus, we present the novel stacked version of our MPN, whose performance can be effectively and continuously improved by stacking multiple submodels.

As an extension of our preliminary paper \cite{Zhang_2019_CVPR}, we propose the event-guided MPN to deal with complex motion blurs in rapidly evolving scenes. To this end, we employ the Dynamic and Active Pixel Sensor (DAVIS) to simultaneously produce the grey-scale Active Pixel Sensor (APS)  and event frames, in which  object motions are captured at a very high temporal resolution ($1\mu s$), thus increasing the ability of our model to deblur complex real-world blurs.

Moreover, we notice that deep deblurring models have a limited robustness to different types of transformations and perturbations, \eg, random rotations, scale transforms, and Gaussian noises. For example, once we apply a weak Gaussian noise to blurry images on input, the PSNR score sharply drops to around 20dB. Therefore, we propose a novel self-supervised robust training strategy to explicitly align deblurred outputs of an input image and its augmented version (deblurred output of augmented image is de-augmented prior to alignment), thus enhancing the robustness of our model. Our contributions are summarized below:
\renewcommand{\labelenumi}{\Roman{enumi}.}
\begin{enumerate}[leftmargin=0.6cm]
\item We propose an end-to-end CNN hierarchical model that performs deblurring in the fine-to-coarse grids by exploiting multi-patch localized-to-coarse operations. Each finer level acts in the residual manner by contributing its residual image to the coarser level, thus letting each level of network focus on different scales of blur. We perform  baseline comparisons in the common testbed (where possible) for  fair comparisons.
\item We identify the limitation to stacking depth of current deep deblurring models and introduce a novel stacking approach which effectively overcomes this limitation.
\item We introduce the use of events in deep multi-patch architecture to capture richer motion information, thus help the model deblur videos with complex blurs and scenes.
\item We propose to apply an auxiliary self-supervised consistency loss leveraging pretext augmentation tasks to enhance the robustness of model \wrt different geometric transformations and photometric distortions, thus reducing overfitting to specific training poses, which helps deblur real-world images.
\end{enumerate}

Our experiments demonstrate clear benefits of our event-guided SPM-like model in non-uniform motion deblurring. To the best of our knowledge, our model is the first multi-patch take on blind motion deblurring, \eg, MPN is the first model that supports deblurring of 720p images real-time (at 30fps). The  self-supervised step is demonstrated useful in deep deblurring scenario also for the first time.

\section{Related Work}
\label{sec:related}
Below we discuss the related works on image deblurring.
Conventional image deblurring methods \cite{cho2009fast,jia2007single,xu2010two,levin2007blind,rajagopalan2014motion,jia2014mathematical,hyun2015generalized,sellent2016stereo} fail to remove non-uniform motion blur due to the use of spatially-invariant deblurring kernel. Moreover, their complex computational inference leads to long processing times, which cannot satisfy the ever-growing needs for real-time deblurring.

\begin{figure*}[htp]
	\centering
	\includegraphics[width=\linewidth]{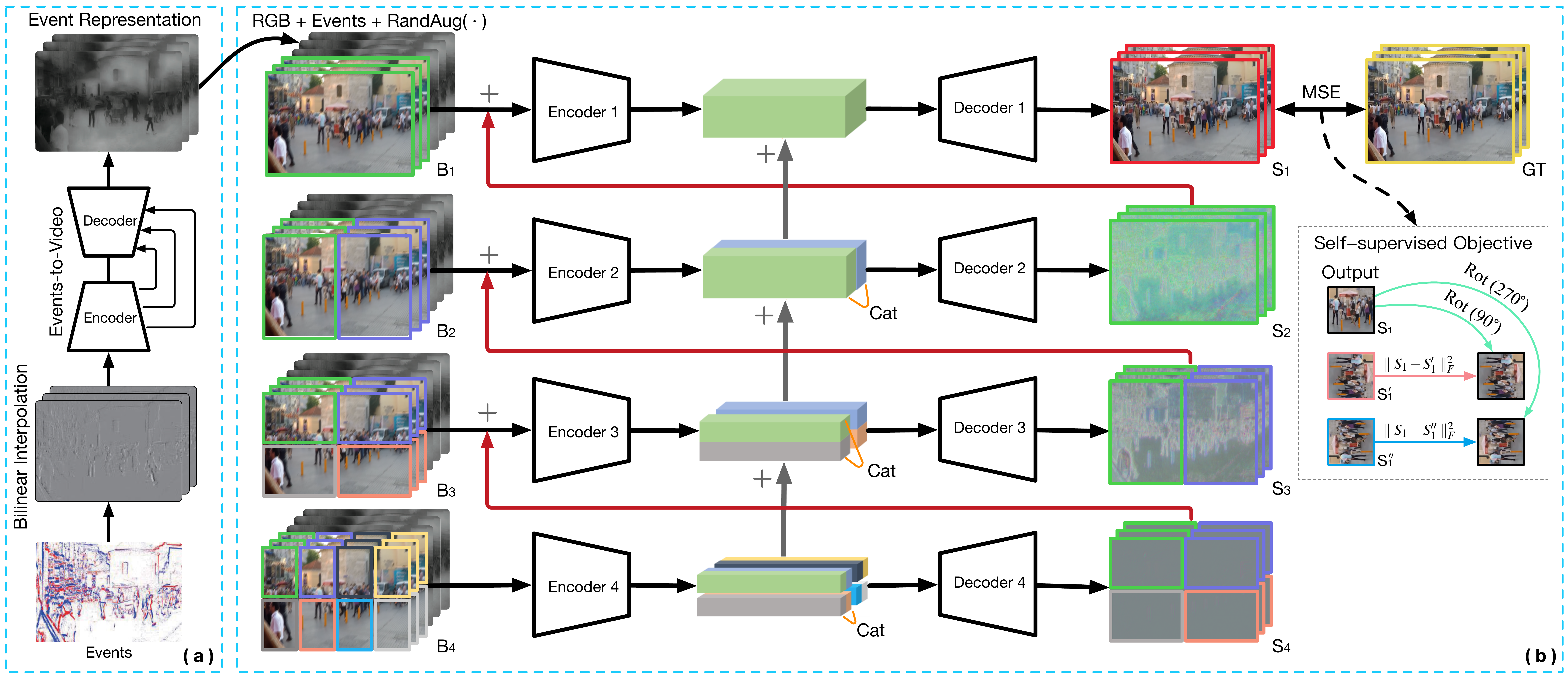}
	\caption{\small Our proposed Event-guided Multi-Patch Network (E-MPN) consists of two parts: (a) a generator of event representation, based on a 10-layer residual U-Net, (b) multi-patch deblurring network which consists of multi-level coarse-to-fine branches. As the patches do not overlap with each other,  they may cause boundary artifacts which are removed by the consecutive upper levels of our model. Symbol $+$ is a summation akin to residual networks.}
	\label{fig:MPN}
\end{figure*}

\vspace{0.05cm}
\noindent\textbf{Deep Deblurring.} Recently, CNNs have been applied for non-uniform image deblurring to deal with the complex motion blur in a time-efficient manner \cite{xu2014deep,sun2015learning,nah2017deep,schuler2016learning,nimisha2017blur,su2017deep}.
Xu \etal \cite{xu2014deep} proposed a deconvolutional CNN which removes blur in non-blind setting by recovering a sharp image given the estimated blur kernel. Their network uses separable kernels which can be decomposed into a small set of filters.  
Sun \etal \cite{sun2015learning} estimated and removed a non-uniform motion blur from an image by learning the regression between 30$\times$30 image patches and their corresponding kernels. The conventional energy-based optimization scheme was employed to estimate the latent sharp image. 

Su \etal \cite{su2017deep} presented a deep learning framework to process blurry video sequences and accumulate information across frames. This method does not require spatially-aligned pairs of samples. 
Nah \etal \cite{nah2017deep} exploited a multi-scale CNN to restore sharp images in an end-to-end fashion from images whose blur is caused by various factors. A multi-scale loss  was employed to mimic the coarse-to-fine pipeline in conventional deblurring approaches.

Recurrent Neural Network (RNN) is often used in deblurring due to its  sequential information processing. Take as an example a network \cite{zhang2018dynamic} consisting of three deep CNNs and one RNN. The RNN is used as a deconvolutional decoder on feature maps extracted by the first CNN module. Another CNN module learns weights for each layer of RNN. The last CNN module reconstructs the sharp image. Scale-Recurrent Network (SRN-DeblurNet) \cite{tao2018scale} uses ConvLSTM cells to aggregate feature maps from coarse-to-fine scales. Finally, Nah \etal \cite{Nah_2019_CVPR} proposed a recurrent network, which iteratively updates the hidden state with existing parameters.

Generative Adversarial Nets (GANs) have also been employed in deblurring due to their advantage in preserving texture details and generating photorealistic images. Kupyn \etal \cite{kupyn2017deblurgan} presented a conditional GAN  which produces high-quality delburred images via the Wasserstein loss.

Gao \etal \cite{Gao_2019_CVPR} proposed a novel selective parameter sharing scheme to improve the dynamic deblurring task. Though their approach achieves impressive results, the complicated nested connections lead to very long processing runtime, which cannot satisfy real-rime applications.

Notably, some recent works have been based on our multi-patch network \cite{Zhang_2019_CVPR}. Suin \etal \cite{suin2020spatially} proposes a modified MPN, which can handle the blur variations across different spatial locations, and adaptively process test images to improve the performance. Dipta \etal \cite{dipta2020fast} propose a fast multi-patch architecture to address image dehazing task.

\noindent\textbf{Event-based modeling.} Events are playing an important role in recent motion deblurring tasks. Event cameras such as DAVIS \cite{davis} and DVS \cite{dvs} record log intensity changes at the microsecond scale with negligible motion blurs, allowing them to compensate the lost information from motion blur. The output of event camera is a stream of events formed into quadruplets $(x,y,t,p)$ that encode the position of brightness changes, time and polarity.

Recent works study how to directly transform events into sharp images, \eg,  Bardow \etal \cite{bardow2016simultaneous} simultaneously estimated the optical flow and intensity images with a fixed-length sliding spatial-temporal window by solving an energy minimizing problem. Barua \etal \cite{barua2016direct} proposed to learn a sparse patch-based dictionary to match event patches with gradient patches, then use the so-called Poison integration to reconstruct the intensity images. Munda \etal \cite{munda2018real} restored intensity images through the manifold regularization. Rebecq \etal \cite{events2video} proposed a novel recurrent network to reconstruct videos from a stream of events. As events are asynchronous, they are raised if there is a local intensity change within the scene, so single events can model static scenes/textures, and sequences of events can model very rapid motions.

DAVIS \cite{davis} can simultaneously output events and Active Pixel Sensor (APS) intensity images that contain the static texture. It directly integrates events on the APS frame and refreshes the event accumulation. Scheerlinck \etal \cite{scheerlinck2018continuous} proposed an asynchronous event-driven complementary filter to integrate the APS frame with events for continuous-time intensity estimation. Pan \etal \cite{pan2019bringing} formulated a deblurring task as an optimization problem that solves a single variable non-convex problem with a double integral model. Jiang \etal \cite{Jiang_2020_CVPR} presented a convolution recurrent network to integrate visual and temporal knowledge at the global and local scales. With a novel directional event filtering module, sharp edge boundary guidance is extracted which increases the quality of reconstructed details. The eSL-Net model \cite{esl} constructed an event-based sparse learning network to improve the deblurring performance. E-CIR \cite{ecir} proposed to leverage events to construct the parametric bases, and introduced a refinement module to propagate visual features among frames. Wang \etal \cite{wang2021asynchronous} proposed to recreate intensity images using an asynchronous Kalman filter based on a unified event and frame uncertainty model. The images reconstructed using these methods, however, include artifacts due to the accumulation of event noises.

\vspace{0.05cm}
\noindent\textbf{Self-supervised learning.} A network can be trained with so-called pretext tasks, \eg, predicting augmentation labels, or predicting easily obtainable self-information as auxiliary objective to improve the performance by making network `aware' of auxiliary tasks. Self-supervised learning has been used in object recognition \cite{dosovitskiy2014discriminative,doersch2015unsupervised,gidaris2018unsupervised,simon2021geodesic,Tas_2021_BMVC}, video representation learning \cite{fernando2017self,sermanet2017time,gan2018geometry,Tas_DA,Wang_2019_ICCV,Wang_2021_ACMMM}, and also few-shot image and video recognition \cite{gidaris2019boosting,su2019boosting,hongguang2020eccv,Zhang_2021_CVPR}. 

Two types of self-supervision are popular: i) contrastive loss; and ii) prediction of label of pretext task. For instance, Gidaris \etal \cite{gidaris2018unsupervised} predict labels of random image rotations, Doersch \etal \cite{doersch2015unsupervised} predict the relative pixel positions, Dosovitskiy \etal \cite{dosovitskiy2014discriminative} learn to discriminate a set of surrogate classes, and  approaches \cite{gidaris2019boosting,su2019boosting} improve the few-shot performance by predicting labels of image rotations and jigsaw patterns. 

In contrast to previous self-supervised pipelines, we leverage self-supervision to promote the consistency of deblurring under augmentations to improve the robustness of model to geometric transformations and photometric distortions. The self-supervision strategy in this paper aligns features obtained from the same augmentation applied at the early and late stage, respectively. This is somewhat related to the so-called knowledge distillation, which encourages one stream of information to distil its knowledge to the other, therefore improving the deblurring performance of our model. However, notice that such a self-supervision is not the distillation pipeline, \ie, it does not use two network streams such as the teacher and student networks.

\begin{figure*}[t]
	\centering
	\includegraphics[width=\linewidth]{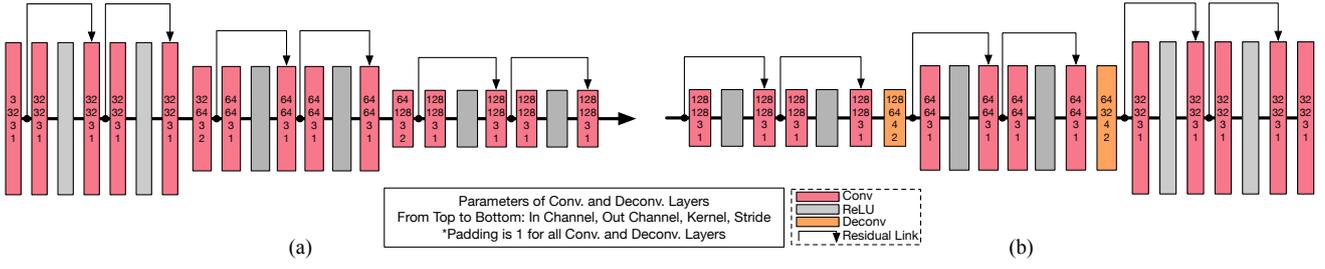}
	\caption{\small The architectures and layer configurations of our (a) decoder and (b) encoder.}
	\label{fig:encoder}
\end{figure*}

\section{Approach}
In this paper, we propose to exploit the multi-patch hierarchy for efficient and effective blind motion deblurring. The overall architecture of our proposed MPN network is shown in Fig.~\ref{fig:MPN} for which we use the (1-2-4-8) model (explained in Sec. \ref{sec:na}) as an example.  
Our network is inspired by coarse-to-fine Spatial Pyramid Matching \cite{lazebnik2006beyond}, which has been used for the problem of  scene recognition \cite{koniusz2018deeper} to aggregate multiple image patches for better performance. In contrast to the expensive inference in multi-scale and scale-recurrent network models \cite{nah2017deep,tao2018scale} shown in Fig.~\ref{fig:Structure_Comp}, our approach (also in Fig.~\ref{fig:Structure_Comp}) uses a residual-like architecture, thus requiring small-size filters which result in fast processing. 
 Despite our model uses a very simple architecture (skip and recurrent connections have been removed), it is very effective. In contrast to Nah \etal \cite{nah2017deep} which uses deconvolution/upsampling links, we use operations such as feature map concatenations, which are possible due to the multi-patch setup we propose. Moreover, our self-supervised unit  differs from typical self-supervised representations: we impose the deblurring consistency between augmented and non-augmented images by reversing the augmentation from the deblurred output.

\subsection{Encoder-decoder Architecture}
Each level of our MPN network consists of one encoder and one decoder whose architecture is illustrated in Fig.~\ref{fig:encoder}.
Our encoder consists of 15 convolutional layers, 6 residual links and 6 ReLU units. The layers of decoder and encoder are identical except that two convolutional layers are replaced by deconvolutional layers to generate images.

Our encoder and decoder use $\sim$3.6 MB parameters due to the small convolutional kernel size and the residual nature of our model, which contribute to the fast deblurring runtime. By contrast, the multi-scale deblurring network \cite{nah2017deep} has 303.6 MB parameters leading to the slower inference.

\begin{figure*}[t]
	\centering
	\includegraphics[width=\linewidth]{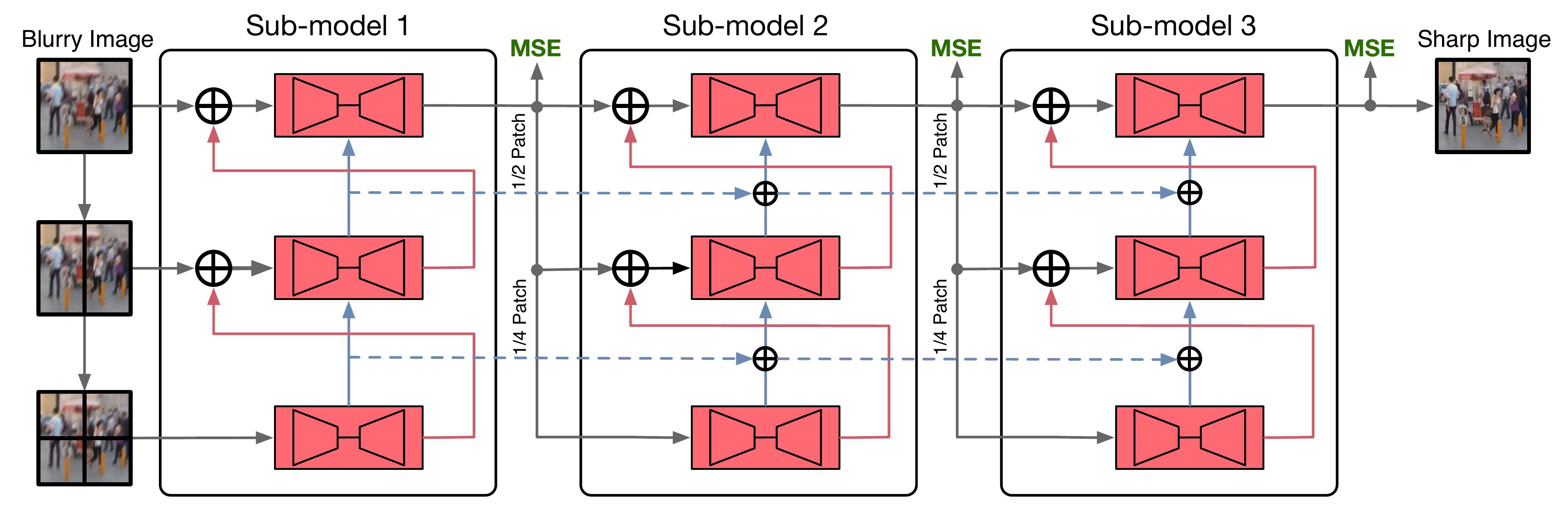} 
	\caption{\small The architecture of Stacked Multi-patch Network (StackMPN), in which multiple MPNs are serially stacked to distribute the deblurring task among several sub-models, and reduce the training difficulty level by level. Such a design provides the performance gain as the  network depth increases.}
	\label{fig:stack}
\end{figure*}

\subsection{Network Architecture}
\label{sec:na}
 Fig.~\ref{fig:MPN} shows the  architecture of our MPN, in which we use the (1-2-4-8) model for illustration purposes.  
Notation (1-2-4-8) indicates the numbers of non-overlapping image patches from the coarsest to the finest level, \ie, a vertical split at the second level, $2\times 2 = 4$ splits at the third level, and $2\times 4 = 8$ splits at the fourth level. For third and fourth levels that output multi-patch residuals, we impose the region-aware consistency loss between adjacent boundaries.  

We denote the initial blurry image input as ${\mathbf B}_1$, while ${\mathbf B}_{i, j}$ is the $j$-th patch at the $i$-th level. Moreover, $\mathcal{F}_i$ and $\mathcal{G}_i$ are the encoder and decoder at level $i$, ${\mathbf C}_{i,j}$ is the output of $\mathcal{G}_i$ for ${\mathbf B}_{i,j}$, and ${\mathbf S}_{i,j}$ represents the output patches from $\mathcal{G}_i$.

Each level of our network consists of an encoder-decoder pair. The input for each level is generated by dividing the original blurry image input ${\mathbf B}_1$ into multiple non-overlapping patches. The output of  encoder from a lower level (corresponds to finer grid) is added to the output of  encoder one level up.
The output of decoder from the lower level (corresponds to finer grid) is added to the upper level input grids passed to the input of  encoder (one level above) so that the top level contains all information inferred in the finer levels. Note that the numbers of input and output patches at each level are different as the main idea of our work is to make the lower level focus on local information (finer grid) to produce a residual information for the coarser gird (obtained by concatenating convolutional features).

Consider the (1-2-4-8) variant as an example. The deblurring process of MPN starts at the bottom level 4. ${\mathbf B}_1$ is sliced into 8 non-overlapping patches ${\mathbf B}_{4,j}, j\!=\!1,\cdots,8$, which are fed into the encoder $\mathcal{F}_4$ to produce the following convolutional feature representation:

\begin{equation}
{\mathbf C}_{4,j} = \mathcal{F}_4({\mathbf B}_{4,j}), \quad j\in\{1,\cdots,8\}.
\end{equation}

Then, we concatenate adjacent features (in the spatial sense) to obtain a new feature representation $C^*_{4,j}$ of the same size as the convolutional feature representation at level 3:

\begin{equation}
{\mathbf C}^*_{4,j} = {\mathbf C}_{4,2j-1} \oplus {\mathbf C}_{4,2j}, \quad j\in\{1,\cdots,4\},
\end{equation}

where $\oplus$ denotes the concatenation operator. The concatenated feature representation ${\mathbf C}^*_{4,j}$ is passed through the encoder $\mathcal{G}_4$ to produce ${\mathbf S}_{4,j} = \mathcal{G}_4({\mathbf C}^*_{4,j})$. 

Next, we move one level up to level 3. The input of $\mathcal{F}_3$ is formed by summing up ${\mathbf S}_{4,j}$ with  patches ${\mathbf B}_{3,j}$. Once the output of $\mathcal{F}_3$ is produced, we add to it ${\mathbf C}^*_{4,j}$:
\begin{equation}
{\mathbf C}_{3,j} = \mathcal{F}_3({\mathbf B}_{3,j} + {\mathbf S}_{4,j}) + {\mathbf C}^*_{4,j}, \quad j\in\{1,\cdots,4\}.
\end{equation}

At level 3, we concatenate the feature representation of level 3 to obtain ${\mathbf C}^*_{3,j}$ and pass it through $\mathcal{G}_3$ to obtain ${\mathbf S}_{3,j}$:
\begin{align}
&{\mathbf C}^*_{3,j} = {\mathbf C}_{3,2j-1} \oplus {\mathbf C}_{3,2j}, \quad j\in\{1, 2\},\\
& {\mathbf S}_{3,j} = \mathcal{G}_3({\mathbf C}^*_{3,j}), \quad j\in\{1,2\}.
\end{align}

Note that features at all levels are concatenated along the spatial dimension: imagine neighboring patches being concatenated to form a larger patch.

At level 2, our network takes two image patches ${\mathbf B}_{2,1}$ and ${\mathbf B}_{2,2}$ as input. We update ${\mathbf B}_{2,j}$ so that ${\mathbf B}_{2,j}\!:=\!{\mathbf B}_{2,j} + {\mathbf S}_{3,j}$ and pass it through $\mathcal{F}_2$:
\begin{align}
&{\mathbf C}_{2,j} = \mathcal{F}_2({\mathbf B}_{2,j} + {\mathbf S}_{3,j}) + {\mathbf C}^*_{3,j}, \quad j\in \{1,2\},\\
&{\mathbf C}^*_2 = {\mathbf C}_{2,1} \oplus {\mathbf C}_{2,2}.
\end{align}

The residual map at level 2 is given by:
\begin{equation}
{\mathbf S}_2 = \mathcal{G}_2({\mathbf C}^*_2).
\end{equation}

At level 1, the final deblurred output $S_1$ is given by:
\begin{align}
&{\mathbf C}_1 = \mathcal{F}_1({\mathbf B}_1 + {\mathbf S}_2) + {\mathbf C}^*_2, \\
&{\mathbf S}_1 = \mathcal{G}_1({\mathbf C}_1).
\end{align}

Different from approaches \cite{nah2017deep,tao2018scale} that evaluate the Mean Square Error (MSE) loss at each level, we evaluate the MSE loss only at the output of level 1 (which resembles the residual network). The loss function of MPN is given as:
\begin{equation}
{\mathcal L}_{deblur} = \frac{1}{2}\sum\limits_{j} \|{\mathbf S}_{1j} - {\mathbf G_j} \|_F^2,
 \label{eq:nostack}
\end{equation}
where $\mathbf{G}_j$ denotes the ground-truth sharp image $j$. 
Due to the hierarchical multi-patch architecture, our network follows the principle of residual learning: the intermediate outputs at different levels ${\mathbf S}_{ij}$ capture  image statistics at different scales. Thus, we evaluate the loss function only at the first level. We have investigated the use of multi-level MSE loss which forces the outputs at each level to be close to the ground truth image. However, as expected, there is no visible performance gain achieved by using the multi-scale MSE loss.

\subsection{Stacked Multi-Patch Network}
As reported by Nah \etal \cite{nah2017deep} and Tao \etal \cite{tao2018scale}, adding finer network levels cannot improve the deblurring performance of the multi-scale and scale-recurrent architectures. For our multi-patch network, we have also observed that dividing the blurred image into ever smaller grids does not further improve the deblurring performance. 
This is mainly due to coarser levels attaining low empirical loss on the training data fast thus excluding the finest levels from contributing their residuals. 

In this section, we propose a novel stacking paradigm for deblurring. Instead of making the network deeper vertically (adding finer levels into the network model, which increases the difficulty for a single worker), we propose to increase the depth horizontally (stacking multiple network models), which employs multiple MPN workers horizontally to perform deblurring. 

Figure \ref{fig:stack} demonstrates how we cascade the MPN to improve the deblurring performance. The stacked model, called StackMPN, stacks multiple ``bottom-top'' MPNs. Note that the output of sub-model $i-1$ and the input of sub-model $i$ are connected, which means that for the optimization of sub-model $i$, output from the sub-model $i-1$ is required. All intermediate features of sub-model $i-1$ are distilled to sub-model $i$. The MSE loss is evaluated at the output of every sub-model $i$ by minimizing the StackMPN objective:  
\begin{equation}
    {\mathcal L}_{deblur} = \frac{1}{2}\sum\limits_{j}\sum\limits_{i=1}^N  \| {\mathbf S}_{ij} - {\mathbf G_j}\|_F^2,
    \label{eq:stack}
\end{equation}

where $N$ is the number of sub-models used, ${\mathbf S}_{ij}$ is the output of sub-model $i$ (note that definitions of ${\mathbf S}_{ij}$ differ between Eq. \ref{eq:nostack} and \ref{eq:stack} ), ${\mathbf G}$ is the ground-truth sharp image, and $j$ loops over a subset of images.

Our experiments will illustrate that such a stacked network can significantly benefit from the increased network depth and improve the deblurring performance accordingly. Although our stacked pipeline uses MPN units, we believe they are generic, that is, other deep deblurring methods can be stacked in the similar manner to improve their performance. However, the total processing time may be unacceptable if a costly deblurring model is employed for the basic unit. Thanks to fast and efficient MPN units, we can control the runtime and size of stacking networks within a reasonable range to cater for various applications. 

\begin{figure*}[t]
	\centering
	\includegraphics[width=0.85\linewidth]{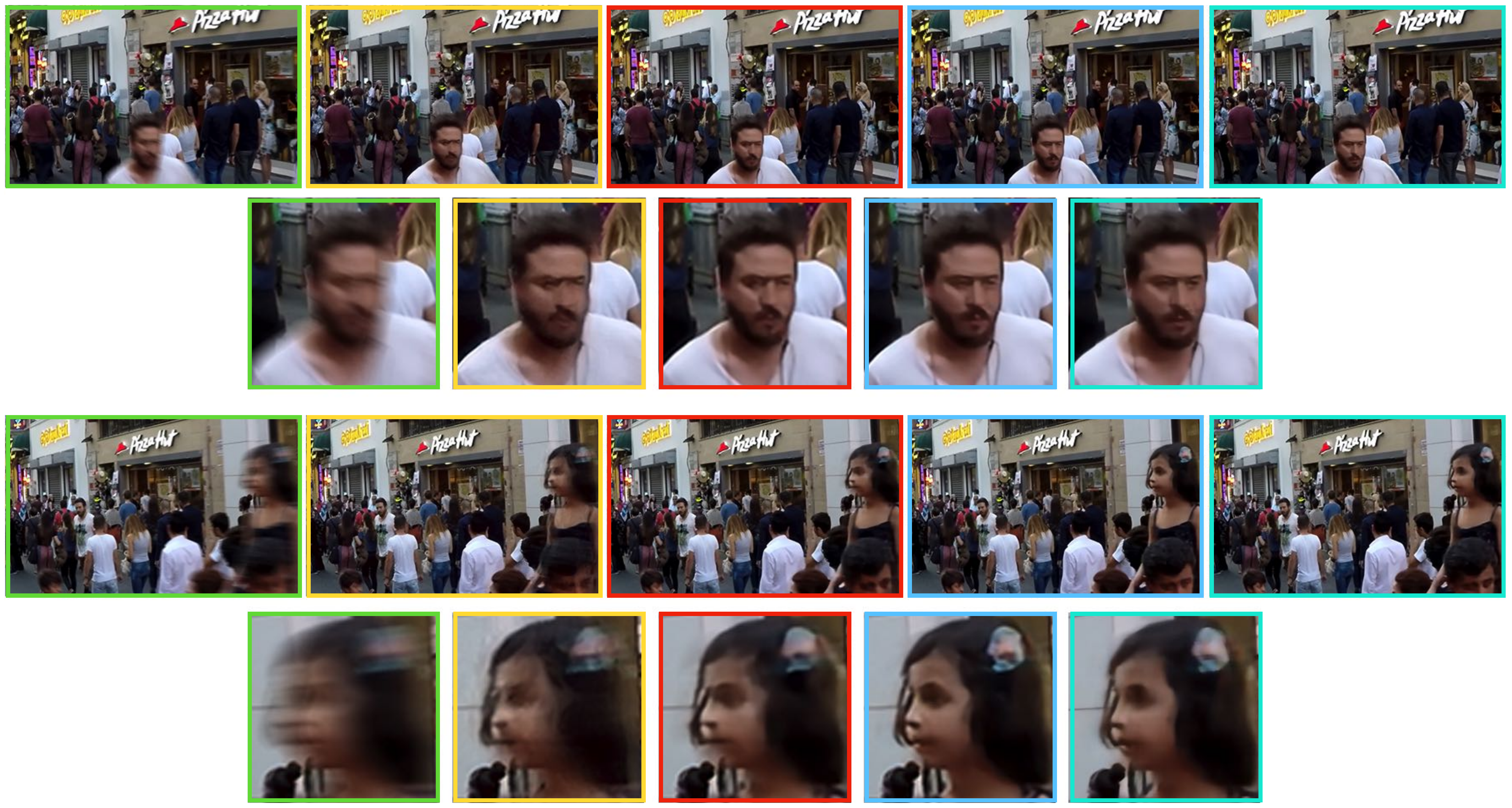}
	\caption{\small Deblurring results. The images from left to right show original blurry images, the results of \cite{nah2017deep}, \cite{tao2018scale}, our MPN and MPN + self-superision, respectively. As can be seen, our method produces the sharpest and most realistic facial details.}
	\label{fig:Comp1}
\end{figure*}

\subsection{Event-guided MPN}
The above proposed MPN model is applied on image deblurring task, thus its ability to deal with realistic and complicated blur patterns is insufficient. To improve deblurring further, one may extend our model to videos in order to  restore the blur kernel based on the temporal information.

One may simply extend MPN by concatenating multiple frames along the channel mode to form an input (followed by the adjustment of the number of input and output channels). However, such a model does not fully exploit the temporal information: the channel-wise convolutional operator in the first layer of encoder  does not guarantee the model to develop a sufficient implicit model of motion. Thus, we propose to introduce the event representation into our MPN  to form a hybrid event-guided deblurring process as in Fig. \ref{fig:MPN}.

Specifically, let us denote $T$ consecutive blurry frames as $\{\mathbf{B}^{(t)}\}_{t=1}^T$,  a set of events as $\{\mathbf{E}^{(t)}\}_{t=1}^T$,  $\Delta\mathbf{E}^{(t)}$ as the event information from time $t$ to $t+\Delta t$, and $\Delta t$, the infinitesimal time step. Let $\{\mathbf{S}^{(t)}\}_{t=1}^T$ be restored sharp frames. Inspired by the design of event cameras \cite{pan2019bringing}, we have:
\comment{
\begin{align}
    \mathbf{B}^{(f)} & = \frac{1}{T}\int_{f-T/2}^{f+T/2}\mathbf{S}^{(t)}\!dt, \label{eq:12}\\
    \mathbf{S}^{(t)} & = \mathbf{S}^{(f)}e^{c\mathbf{E}^{(t)}}, \label{eq:13}\\
    \mathbf{E}^{(t)} & = \int_f^t\Delta\mathbf{E}^{(h)}dh, \label{eq:14}
\end{align}}

\begin{align}
    \mathbf{B}^{(f)} & = \frac{1}{T}\sum\limits_{t = f-T/2}^{f+T/2}\mathbf{S}^{(t)}, \label{eq:12}\\
    \mathbf{S}^{(t)} & = \mathbf{S}^{(f)}e^{c\mathbf{E}^{(t)}}, \label{eq:13}\\
    \mathbf{E}^{(t)} & = \sum\limits_{h=f}^t\Delta\mathbf{E}^{(h)}, \label{eq:14}
\end{align}

where $c$ is the threshold determining if an event should be recorded, $\mathbf{E}^{(t)}$ is the sum of events during time $f$ to $t$.

Above equations show that the blurry frame is generated from latent sharp frames during the exposure time $[f-T/2,f+T/2]$, and the sharp frame at time $t$ can be generated by simply interpolating the sharp frame at time $t-T$ and the events during time step $T$.   
Combining Eq. \ref{eq:12}, \ref{eq:13} and \ref{eq:14} , we have:
\comment{
\begin{align}
    & \mathbf{B}^{(f)} = \frac{1}{T}\int_{f-T/2}^{f+T/2}\mathbf{S}^{(t)}\!dt = \frac{\mathbf{S}^{(f)}}{T}\int_{f-T/2}^{f+T/2}e^{c\int_f^t\Delta\mathbf{E}^{(h)}dh} dt, \\
    & \log \mathbf{S}^{(f)}  = \log\mathbf{B}^{(f)} - \log\left(\frac{1}{T}\int_{f-T/2}^{f+T/2}e^{c\,\mathbf{E}^{(t)}}\!dt\right).
\end{align}}

\begin{align}
    & \mathbf{B}^{(f)} = \frac{1}{T}\sum\limits_{t=f-T/2}^{f+T/2}\mathbf{S}^{(t)}
    = \frac{\mathbf{S}^{(f)}}{T}\sum\limits_{t=f-T/2}^{f+T/2}e^{c\sum_{h=f}^t\Delta\mathbf{E}^{(h)}}, \\
    & \log \mathbf{S}^{(f)}  = \log\mathbf{B}^{(f)} - \log\left(\frac{1}{T}\sum\limits_{t=f-T/2}^{f+T/2}e^{c\,\mathbf{E}^{(t)}}\right).
\end{align}

From the above equations, we conclude that the sharp frame is associated with the original blurry frame and the event information. Once the event information from $f-T/2$ to $f+T/2$ is collected, the learning formulation of MPN can be re-written as the following two-stream variant:

\begin{align}
    \mathbf{S}^{(f)}&  = \text{MPN}(B^{(f)}, \; \mathbf{E}^{(f-T/2:f+T/2)}).
\end{align}

We evaluate our model on realistic event datasets \cite{pan2019bringing} in which samples are recorded by DAVIS sensor. For non-event datasets, we employ the method proposed in \cite{esim} to simulate the event information from RGB frames.

The event stream needs to be converted to an image-like  event representation before being fed into the pipeline for training and evaluation. Integrating the event on a 2D plane is a natural choice, whereas encoding it with spatial-temporal voxel grad can be more effective. Thus, we use the Events-to-Video model \cite{events2video}, which is built upon a 10-layer residual U-Net, to produce 10 adjacent event representations with $0.1s$ time grid for the central frame. Subsequently, 10 event representations are concatenated with original RGB blurry frames to restore the sharp RGB output. As the events are collected at a very high frame-rate, we naturally capture accurate object motions compared to conventional cameras. Thus, our two-stream deblurring model is expected to significantly improve the deblurring performance in end-to-end manner for both synthetic and realistic blurry images.

\begin{figure*}[t]
    \centering
    \includegraphics[width=\linewidth]{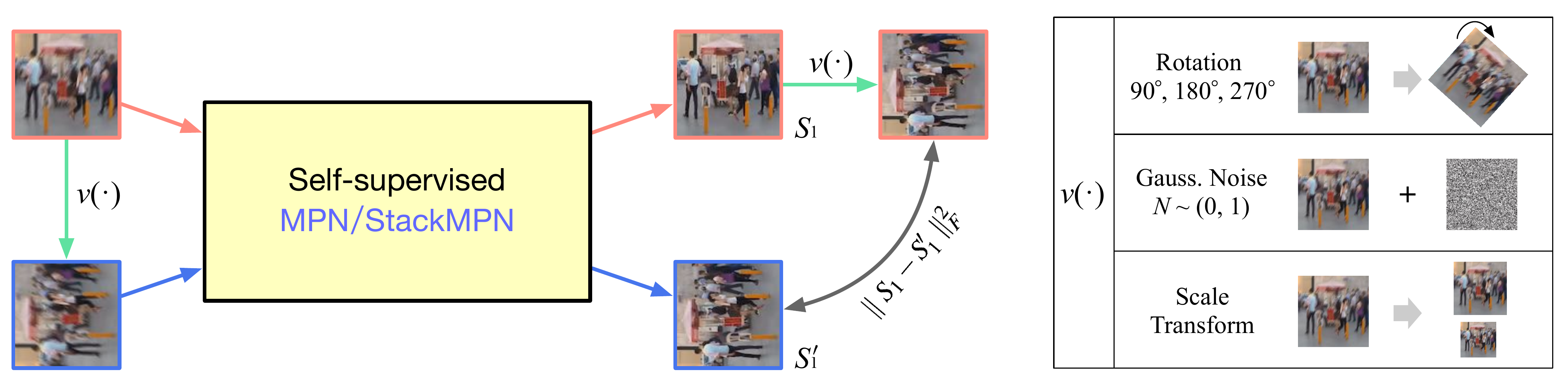}
    \caption{\hg{Self-supervised training step. The consistency loss is applied between the augmented deblurred images of vanilla images and the deblurred outputs of augmented counterparts, thus promoting the robustness to various transformations and noises during training and inference. Functions $v(\cdot)$ denote a chosen augmentation. Note that the self-supervised loss does not enforce the ground truth on the output which limits overfitting.}}
    \label{fig:ar-mpn}
\end{figure*}

\subsection{Boosting MPN with Self-supervision}
Below we introduce the self-supervision to boost the performance and robustness of our MPN approach. Figure \ref{fig:ar-mpn} illustrates the self-supervised aspect of our pipeline, for which we investigate the impact of rotations, scale transform and Gaussian noise augmentations. 

\vspace{0.1cm}
\noindent\textbf{Rotations.} One natural property of blur kernels is the invariance to rotations, which  enhances the robustness and generalization ability in realistic scenarios. However, previous works \cite{nah2017deep,tao2018scale,zhang2018dynamic} do not guarantee such a property even though they use the random rotation augmentations. Thus, we  propose to improve the robustness of MPN to rotations by promoting the consistency between original and rotated restored outputs in a self-supervised manner. 

Let $\mathbf{B}_j$ be a blurred image, $\text{Deblur}(\cdot)$ be a deblurring network, \eg, MPN or StackMPN, $\text{Rot}(\cdot)$ be the rotation function (with random choice of rotation by $90^\circ, 180^\circ$, or $270^\circ$). The self-supervised loss $L_{ss}$ is defined as:

\begin{align}
\centering
    \mathbf{S}_{Nj} &= \text{Deblur}(\mathbf{B}_j), \\
    \mathbf{S}'_{Nj} &= \text{Deblur}({v}(\mathbf{B}_j)),\\
    {\mathcal L}_{ss} &= \sum_j|| {v}(\mathbf{S}_{Nj}) - \mathbf{S}'_{Nj} ||_F^2,
    \label{eq:rot}
\end{align}

where $j$ loops over a subset of images, $N$ is the number of stacked levels of network, $\text{Deblur}(\cdot)$ is the output from level $N$ of stacked network, whereas $v(\cdot)$ performs a chosen augmentation, \eg, $\text{Rot}(\cdot)$.

In this manner, the deblurring network is exposed to a varienty of orientations and can capture the rotation-invariant blur kernels. With the rotation-based self-supervised loss term, our final objective of MPN is defined as follows:
\begin{equation}
    {\mathcal L} = {\mathcal L}_{deblur} + \alpha {\mathcal L}_{ss},
    \label{eq:tot_l}
\end{equation}
where $\alpha\geq 0$ is the hyper-parameter to tune.

\vspace{0.1cm}
\noindent\textbf{Scale Transformations.} Another well-established property of blurry kernels is their consistency for different scale inputs. Recent works \cite{nah2017deep,zhang2018dynamic} exploit the multi-scale inputs for image deblurring. As discussed in Section \ref{sec:related}, the improvement from such a design is not significant compared to our multi-patch network. Ideally, one might replace each level of MPN with a multi-scale architecture, thus making it a multi-scale multi-patch network, denoted as `MPN+MSN' in Table \ref{tabel:PSNR_sc}, to simultaneously capture non-uniform kernels from different scales and locations. However, such a design is extremely costly \wrt network parameters, training overheads and inference time, thus not practical.

Instead, we follow the self-supervision step designed in a similar spirit to the rotation-based self-supervision step, with the goal of promoting the consistency between scales, and capturing multi-scale information in an efficient self-supervised manner. The difference compared to the rotation-based self-supervision is that instead  of batch of  rotated images, we randomly downsample or upsample the original blurry images, and simultaneously feed the original input and images at different scales to MPN to obtain their outputs. We follow Eq. \ref{eq:rot} and \ref{eq:tot_l} but simply use a scale augmenting function $\text{Scale}(\cdot)$ in the place of $\text{Rot}(\cdot)$.

\begin{table}[t]
\centering
\caption{\small Quantitative analysis of our model on the GoPro dataset \cite{nah2017deep}. {\em Size} and {\em Runtime} are expressed in MB and milliseconds. The reported time is the CNN runtime (writing generated images to disk is not considered). Note that we employ (1-2-4) multi-patch architecture for StackMPN and E-StackMPN. We did not investigate deeper stacking networks due to the GPU memory limits and long training times.}
\makebox{\begin{tabular}{l|c|c|c|c}
\hline
Models &  PSNR & SSIM & Size & Runtime\\ \hline
{\color{pink}$^\clubsuit$}Sun \etal \cite{sun2015learning} & 24.64 & 0.843 & 54.1 & {\small 12000} \\
{\color{pink}$^\clubsuit$}Nah \etal \cite{nah2017deep} & 29.23 & 0.916 & 303.6 & {\small 4300}\\
{\color{pink}$^\clubsuit$}Zhang \etal \cite{zhang2018dynamic} & 29.19 & 0.931 & 37.1 & {\small 1400}\\
{\color{pink}$^\clubsuit$}Tao \etal \cite{tao2018scale} & 30.10 & 0.932 & 33.6 & {\small 1600}\\ 
{\color{pink}$^\clubsuit$}Gao \etal \cite{Gao_2019_CVPR} & 30.92 & 0.942 & 2.8 & {\small N/A} \\ 
{\color{pink}$^\clubsuit$}Park \etal \cite{park2020multi} & 31.15 & 0.945 & 2.6 & 70 \\
{\color{lblue}$^\blacklozenge$}Nah \etal \cite{Nah_2019_CVPR} & 29.97 & 0.895 & {\small N/A} & {\small 34.7} \\
{\color{lblue}$^\blacklozenge$}DBGAN \cite{DBGAN} & 31.10 & 0.943 & 11.8 & {\small N/A} \\ 
{\color{tiffany}$^\spadesuit$}Pan \etal \cite{pan2019bringing} & {29.06} & {0.943} & {\small N/A} & {\small N/A} \\
{\color{tiffany}$^\spadesuit$}eSL-Net \cite{esl} & 30.23 & 0.870 & {\small N/A} & {\small N/A} \\
{\color{tiffany}$^\spadesuit$}Jiang \etal \cite{Jiang_2020_CVPR} & 31.79 & 0.949 & {\small N/A} & {\small N/A} \\
{\color{tiffany}$^\spadesuit$}Pan \etal \cite{pan2020cascaded} & 31.89 & 0.921 & {\small 286} & {\small 6500} \\
{\color{tiffany}$^\spadesuit$}Xiang \etal \cite{xiang2020deep} & 32.63 & 0.935 &{\small 61.8} &{\small 6000}  \\
\hline
\multicolumn{5}{c}{(@ indicates the number of RGB frames used)} \\ \hline
{\color{pink}$^\clubsuit$}MPN & 30.21 & 0.935 & 21.7 & 17 \\
{\color{pink}$^\clubsuit$}StackMPN & 31.16 & {0.945} & 65.1 & 233 \\
\arrayrulecolor{black} \cdashline{1-5}[1pt/3pt]
{\color{lblue}$^\blacklozenge$}MPN @3 & 30.68 & 0.940 & \multirow{3}{*}{21.7} & \multirow{3}{*}{17}\\
{\color{lblue}$^\blacklozenge$}MPN @5 & 30.89 & 0.941 &  & \\
{\color{lblue}$^\blacklozenge$}MPN @7 & 30.58 & 0.939 &  & \\
{$^\blacklozenge$}StackMPN @5 & 31.63 & \textbf{0.951} & 65.1 & 233\\
\arrayrulecolor{black} \cdashline{1-5}[1pt/3pt]
{\color{tiffany}$^\spadesuit$}E-MPN & 33.14 & 0.937 & 64.7 & 121 \\
{\color{tiffany}$^\spadesuit$}E-MPN @5 & 33.24 & 0.936 & 64.7 & 121 \\
{\color{tiffany}$^\spadesuit$}E-StackMPN & {33.56} & 0.939 & 118.2 & 338 \\
{\color{tiffany}$^\spadesuit$}E-StackMPN @5 & \textbf{33.83} & 0.941 & 118.2 & 338 \\
\hline
\multicolumn{5}{l}{{\color{pink}$\clubsuit$}: single image deblur; {\color{lblue}$\blacklozenge$}: video deblur; {\color{tiffany}$\spadesuit$}:event-based deblur.}\\
\end{tabular}}
\label{tabel:PSNR}
\vspace{-0.2cm}
\end{table}

\begin{table}[t]
    \centering
    \caption{Ablations on the performance of hierarchical architecture.}
    \begin{tabular}{l|c|c|c|c}
    \hline
    Models & \small PSNR & \small SSIM & Size & \small Runtime\\ \hline
    \small MPN(1) & 28.70 &  0.9131 & 7.2 & 5\\
    \small MPN(1-2) & 29.77 & 0.9286 & 14.5 & 9\\
    \small MPN(1-1-1) & 28.11 & 0.9041 & 21.7 & 12\\
    \small MPN(1-2-4) & 30.21 & 0.9345 & 21.7 & 17\\
    \small MPN(1-4-16) & 29.15 & 0.9217 & 21.7 & 92\\
    \small MPN(1-2-4-8) & \textbf{30.25} & \textbf{0.9351} & 29.0 & 30\\
    \small MPN(1-2-4-8-16) & 29.87 & 0.9305 & 36.2 & 101 \\ \hline
    \end{tabular}
    \label{tab:my_label}
\end{table}

\vspace{0.1cm}
\noindent\textbf{Gaussian Noise.} The robustness to an additive pixel noise drawn from the Normal distribution is a desired property  to equip a deblurring model with. Previous works \cite{nah2017deep,tao2018scale,zhang2018dynamic}, including our MPN, have no built-in robustness to Gaussian noises. To demonstrate this point, Table \ref{tabel:robust-comp} shows that the inference performance sharply decreases once small-valued noises are injected into the blurring test images. Previous works randomly apply noises to blurred images and use them during training. Thus, the model is expected to deal with deblurring and denoising simultaneously.

\section{Experiments}
\label{sec:expt}
Below, we present experimental evaluations of several variants of MPN. Firstly, we introduce datasets we use.
\subsection{Datasets}
We train/evaluate our methods on several versions of the GoPro dataset \cite{nah2017deep} and the VideoDeblurring dataset \cite{su2017deep}, and perform qualitative analysis on the realistic blurry images \cite{pan2019bringing} to visually compare the deblurring ability of each model. Lastly, we capture some realistic heavily blurred images consisting of both camera motion and object motion to further justify real-life the effectiveness of each method.

\vspace{0.1cm}
\noindent{\textbf{GoPro}} dataset \cite{nah2017deep} consists of 3214 pairs of blurred and clean images extracted from 33 sequences at 720$\times$1280 resolution. The blurred images  are generated by averaging varying number (7--13) of successive latent frames to produce varied blurs. For a fair comparison, we follow the protocol in \cite{nah2017deep}, which uses 2103 image pairs for training and the remaining 1111 pairs for testing.

\vspace{0.1cm}
\noindent{\textbf{VideoDeblurring}} dataset \cite{su2017deep} contains videos captured by various devices, such as iPhone, GoPro and Nexus. The quantitative part has 71 videos. Every video consists of 100 frames at 720$\times$1280 resolution. Following the setup in \cite{su2017deep}, we use 61 videos for training and the remaining 10 videos for testing. In addition, we evaluate the model trained on the GoPro dataset \cite{nah2017deep} on the VideoDeblurring dataset to demonstrate the generalization ability of our method.

\subsection{Evaluation Setup and Results}
We feed the original high-resolution $720\!\times\!1280$ pixel images into MPN.
The PSNR, SSIM, model size and runtime are reported in Table \ref{tabel:PSNR} for an in-depth comparison with competing state-of-the-art motion deblurring models. For the stacking networks, we employ the (1-2-4) multi-patch architecture in every model unit,  considering the runtime and difficulty of training.

\begin{table}[t]
\centering
\caption{\small The baseline performance of multi-scale and multi-patch methods on the GoPro dataset \cite{nah2017deep}. MSN is a baseline that uses our encoder and decoder following the design of \cite{nah2017deep} (refer to baselines from Section \ref{sec:expt} for details). 
Note that MSN(1) and MPN(1) are in fact the same model.
} 
\makebox{\begin{tabular}{l|c|c|c}
\hline
Models & PSNR & SSIM & Runtime\\ \hline
Nah \etal \cite{nah2017deep} & 29.23 & 0.9162 & 4300\\ \hline
MSN(1) &  \multirow{2}{*}{28.70} & \multirow{2}{*}{0.9131} & \multirow{2}{*}{4}\\ 
MPN(1) &    &  & \\
\arrayrulecolor{black} \cdashline{1-4}[1pt/3pt]
MSN(2) &  28.82 & 0.9156 & 21\\
MPN(1-2) & 29.77 & 0.9286 & 9 \\
\arrayrulecolor{black} \cdashline{1-4}[1pt/3pt]
MSN(3) &  28.97 &  0.9178 & 27\\
MPN(1-2-4) & 30.21 & 0.9345 & 17\\
\arrayrulecolor{black} \cdashline{1-4}[1pt/3pt]
MPN + MSN & 30.34 & 0.9351 & 523 \\ \hline
\end{tabular}}
\label{tabel:PSNR_sc}
\end{table}

\begin{table*}[t]
\centering
\caption{\small Quantitative analysis (PSNR) on the VideoDeblurring dataset \cite{su2017deep} for models trained on the GoPro dataset. PSDeblur means using Photoshop CC 2015.
We select the ``single frame'' version of approach \cite{su2017deep} for fair comparisons.}
\makebox{\begin{tabular}{l|c|c|c|c|c|c|c|c|c|c|c}
\hline
Methods & \#1 & \#2 & \#3 & \#4 & \#5 & \#6 & \#7 & \#8 & \#9 & \#10 & {\small Average} \\ \hline
{\color{pink}$^\clubsuit$}PS\small{Deblur} \cite{su2017deep}  &24.42 & 28.77 & 25.15 & 27.77 & 22.02 & 25.74 & 26.11 & 19.75 & 26.48 & 24.62 & 25.08 \\
{\color{pink}$^\clubsuit$}WFA \cite{delbracio2015hand} & 25.89 & 32.33 & 28.97 & 28.36 & 23.99 & 31.09 & 28.58 & 24.78 & 31.30 & 28.20 & 28.35 \\
{\color{pink}$^\clubsuit$}Su \etal \cite{su2017deep} & 25.75 & 31.15 & 29.30 & 28.38 & 23.63 & 30.70 & 29.23 & 25.62 & 31.92 & 28.06 & 28.37 \\
{\color{lblue}$^\blacklozenge$}Nah \etal \cite{Nah_2019_CVPR} & - & - & - & - & - & - & - & - & - & - & 30.80 \\ 
{\color{lblue}$^\blacklozenge$}STFAN \cite{zhou2019spatio} & - & - & - & - & - & - & - & - & - & - & 31.24 \\
{\color{lblue}$^\blacklozenge$}Xiang \etal \cite{xiang2020deep} & - & - & - & - & - & - & - & - & - & - & 31.68 \\ 
{\color{lblue}$^\blacklozenge$}Pan \etal \cite{pan2020cascaded} & - & - & - & - & - & - & - & - & - & - & 31.67 \\
\hline
{\color{pink}$^\clubsuit$}MPN & 29.89 & 33.35 & 31.82 & 31.32 & 26.35 & 32.49 & 30.51 & 27.11 & 34.77 & 30.02 & 30.76 \\
{\color{pink}$^\clubsuit$}StackMPN & 30.48 & 34.31 & 32.24 & 32.09 & 26.77 & 33.08 & 30.84 & 27.51 & 35.24 & 30.57 & 31.39 \\
\arrayrulecolor{black} \cdashline{1-12}[1pt/3pt]
{\color{tiffany}$^\spadesuit$}E-MPN & {31.32} & {34.73} & {33.21} & {32.68} & {27.85} & {33.81} & {31.83} & {28.46} & {36.09} & {31.45} & {32.14} \\
{\color{tiffany}$^\spadesuit$}E-StackMPN & \textbf{31.68} & \textbf{35.81} & \textbf{33.35} & \textbf{33.17} & \textbf{28.32} & \textbf{34.17} & \textbf{32.15} & \textbf{29.01} & \textbf{36.27} & \textbf{31.75} & \textbf{32.57} \\ \hline
\multicolumn{5}{l}{{\color{pink}$\clubsuit$}: single image deblur; {\color{lblue}$\blacklozenge$}: video deblur; {\color{tiffany}$\spadesuit$}:event-based deblur.}\\
\end{tabular}}
\label{tabel:videodeblurringresults}
\end{table*}

\begin{figure}[b]
	\centering
	\includegraphics[width=\linewidth]{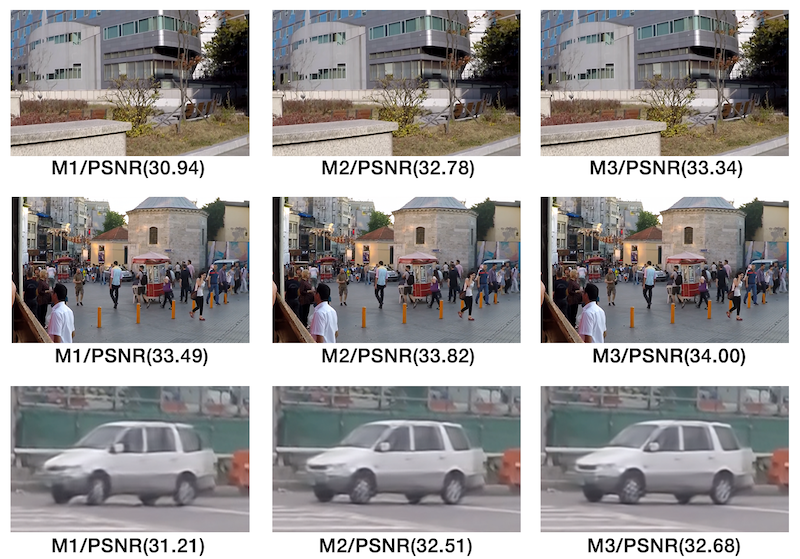}
    \caption{\small Outputs of different sub-models of Stack(3)-MPN. From left to right are the outputs of ${\mathbf M}_1$ to ${\mathbf M}_3$. The clarity of results improves level-by-level. }
	\label{fig:sdnet_comp}
\end{figure}

For the stacked model, the output of every sub-model is optimized level-by-level, which means the first output has the poorest quality and the last output achieves the best performance. Fig.~\ref{fig:sdnet_comp} presents the outputs of Stack(3)-MPN (3 sub-models stacked together) to demonstrate that each sub-model gradually improves the quality of deblurring.

\subsection{Implementation Details}
All our experiments are implemented in PyTorch and evaluated on a single NVIDIA Tesla P100. To train MPN, we randomly crop images to $256\!\times\!256$ pixel size. Subsequently, we extract patches from the cropped images  and forward them to the inputs of each level. The batch size is set to 6 during training. The Adam solver \cite{kingma2014adam} is used to train our models for 3000 epochs. The initial learning rate is set to 0.0001 and the decay rate to 0.1. We normalize image to range $[0, 1]$ and subtract 0.5.

\begin{table}[t]
\centering
\caption{\small Evaluations of the weight sharing scheme on GoPro \cite{nah2017deep}.}
\makebox{\begin{tabular}{l|c|c|c}
\hline
Models & PSNR & SSIM & Size (MB)\\ \hline
MPN(1-2) & 29.77 & 0.9286 & 14.5\\
MPN(1-2)-WS & 29.22 & 0.9210 & 7.2\\
\hdashline
MPN(1-2-4) & 30.21 & 0.9343 & 21.7\\
MPN(1-2-4)-WS & 29.56 & 0.9257 & 7.2\\
\hdashline
MPN(1-2-4-8) & 30.25 & 0.9351 & 29.0\\
MPN(1-2-4-8)-WS & 30.04 & 0.9318 & 7.2\\ \hline
\end{tabular}}
\label{tabel:share_weight}
\end{table}

\comment{
\begin{table}[t]
\centering 
\caption{\small Ablation study of the robustness of  MPN \vs MPN+self-supervision under different transformations on GoPro \cite{nah2017deep}. The robustness to augmentations is measured by randomly applying transformations or noises on test images to measure the PSNR.}
\vspace{0.2cm}
\makebox{\begin{tabular}{l|c|c|c|c}
\hline
Augmentations & MPN & MPN* & SsR-MPN & SsR-MPN* \\ \hline
Rotation & \multirow{3}{*}{30.24} & 29.89 & 30.85 & 30.86\\
Gaussian Noise &  & 20.98 & 30.46 & 30.31 \\
Scale &  & 24.71 & 30.35 & 29.49  \\ \hline
\multicolumn{5}{l}{*: applying augmentations on test samples.} 
\end{tabular}}
\label{tabel:robust-comp}
\end{table}}

\begin{table}[t]
\centering 
\caption{\small Ablation study of the robustness of  MPN \vs MPN  models with self-supervision under different transforms/noises on  GoPro \cite{nah2017deep}. The robustness to augmentations is measured by randomly applying transformations or noises on test images to measure the PSNR.}
\makebox{\begin{tabular}{l|c|c|c|c}
\hline
Type of Aug. & None & Rot. & Gauss. & Scale \\ \hline
MPN & 30.24 & 29.89 & 20.98 & 24.71\\
MPN + Rand. Aug. & 30.01 & 29.90 & 28.05 & 28.31  \\ 
\hdashline
MPN + SS (Rot.) & 30.85 & 30.86 & - & - \\
MPN + SS (Gauss.) & 30.46 & - & 30.31 & -   \\ 
MPN + SS (Scale) & 30.35 & - & - & 29.49 \\ 
MPN + SS (Mix) & 30.92 & 30.91 & 30.35 & 29.91 \\ \hline
\end{tabular}}
\label{tabel:robust-comp}
\end{table}

\begin{figure*}[t]
    \centering
    \includegraphics[width=\linewidth]{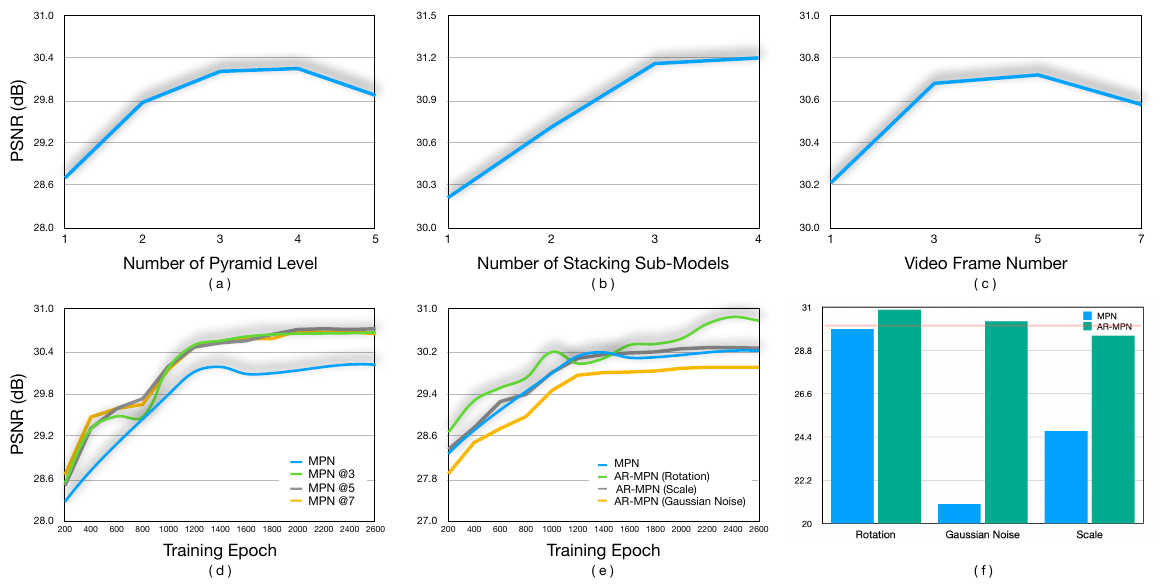}
    \caption{Ablation studies. PSNR \wrt ({\em a}) the number of pyramid levels, ({\em b}) stacking units, ({\em c}) the number of frames concatenated. Moreover, we show PSNR \wrt to ({\em d}) the number of video frames used (indicated by @) by MPN, and ({\em e}) the type of augmentations as a function of epoch number. Finally, ({\em f}) compares the final performance of MPN alone \vs MPN with different self-supervision strategies.}
    \label{fig:comp_3}
    \vspace{-0.4cm}
\end{figure*}
\comment{
\begin{figure*}[t]
    \centering
    \includegraphics[width=\linewidth]{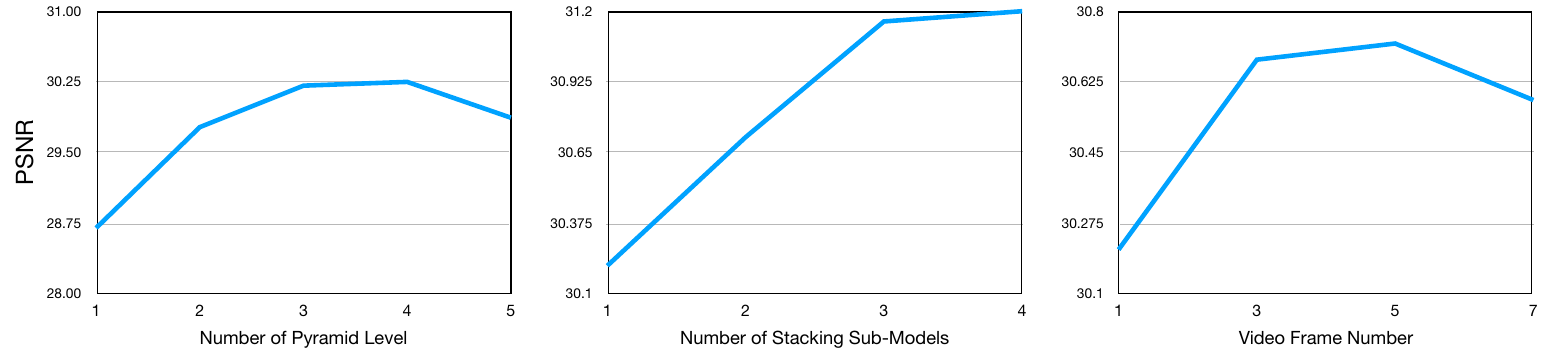}
    \caption{The ablations \textit{w.r.t} the number of pyramid level, }
    \label{fig:training-curve}
\end{figure*}

\begin{figure}[t]
    \centering
    \includegraphics[width=\linewidth]{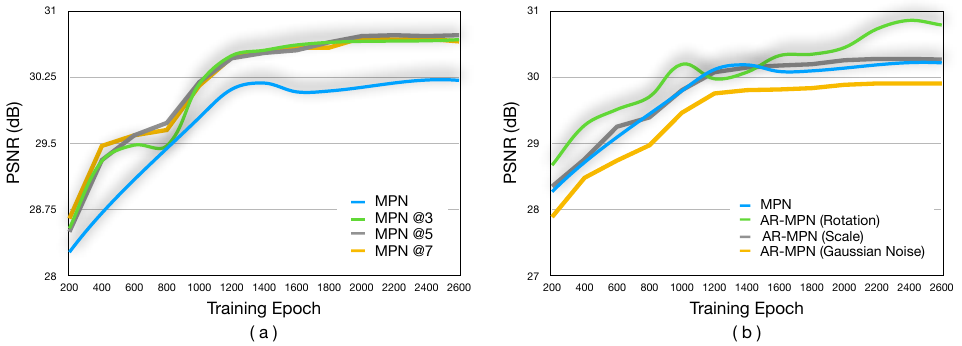}
    \caption{The PSNR-Epoch curves of our proposed models. (a) Comparisons between image and video deblurring curves with different frame numbers. (b) Comparisons between original MPN and the three Self-supervised Robust MPNs (SsR-MPN).}
    \label{fig:training-curve}
\end{figure}
}

\noindent{\textbf{Performance.}} Table \ref{tabel:PSNR} shows that our proposed MPN outperforms other competing methods according to PSNR and SSIM measures, which demonstrates the superiority of non-uniform blur removal via the localized information our model uses. The deepest MPN we trained and evaluated is (1-2-4-8-16) due to the GPU memory limitation. The best performance is obtained with the (1-2-4-8) model, for which PSNR and SSIM are higher compared to all current state-of-the-art models. Note that our model is simpler than other competing approaches, \eg, we do not use recurrent units. We note that patches that are overly small (below 1/16 size) are not helpful in removing the motion blur. 

Moreover, the stacked variant, StackMPN, outperforms shallower MPN by around 1.0dB PSNR. SSIM scores indicate the same trend. The performance of StackMPN can be improved by serially stacking more MPN units, which is consistent with our expectations.

\begin{figure}[t]
	\centering
	\includegraphics[width=\linewidth]{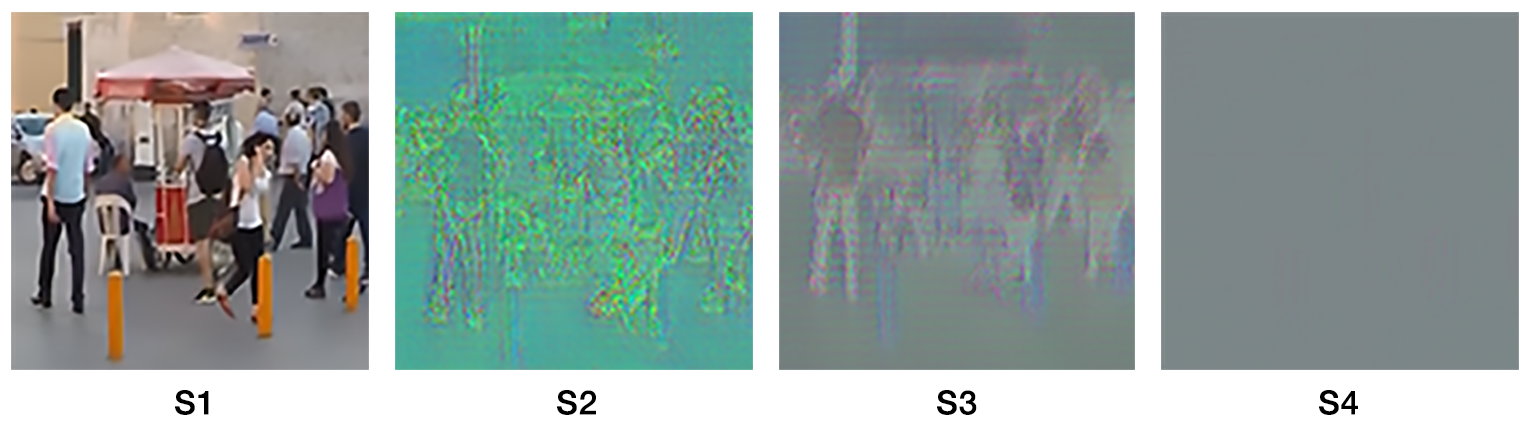}
    \caption{\small Outputs ${\mathbf S}_i$ for different levels of MPN(1-2-4-8). Images from right to left visualize bottom level ${\mathbf S}_4$ to top level ${\mathbf S}_1$.}
	\label{fig:dmphn_comp}
\end{figure}

\begin{figure}
    \centering
    \includegraphics[width=\linewidth]{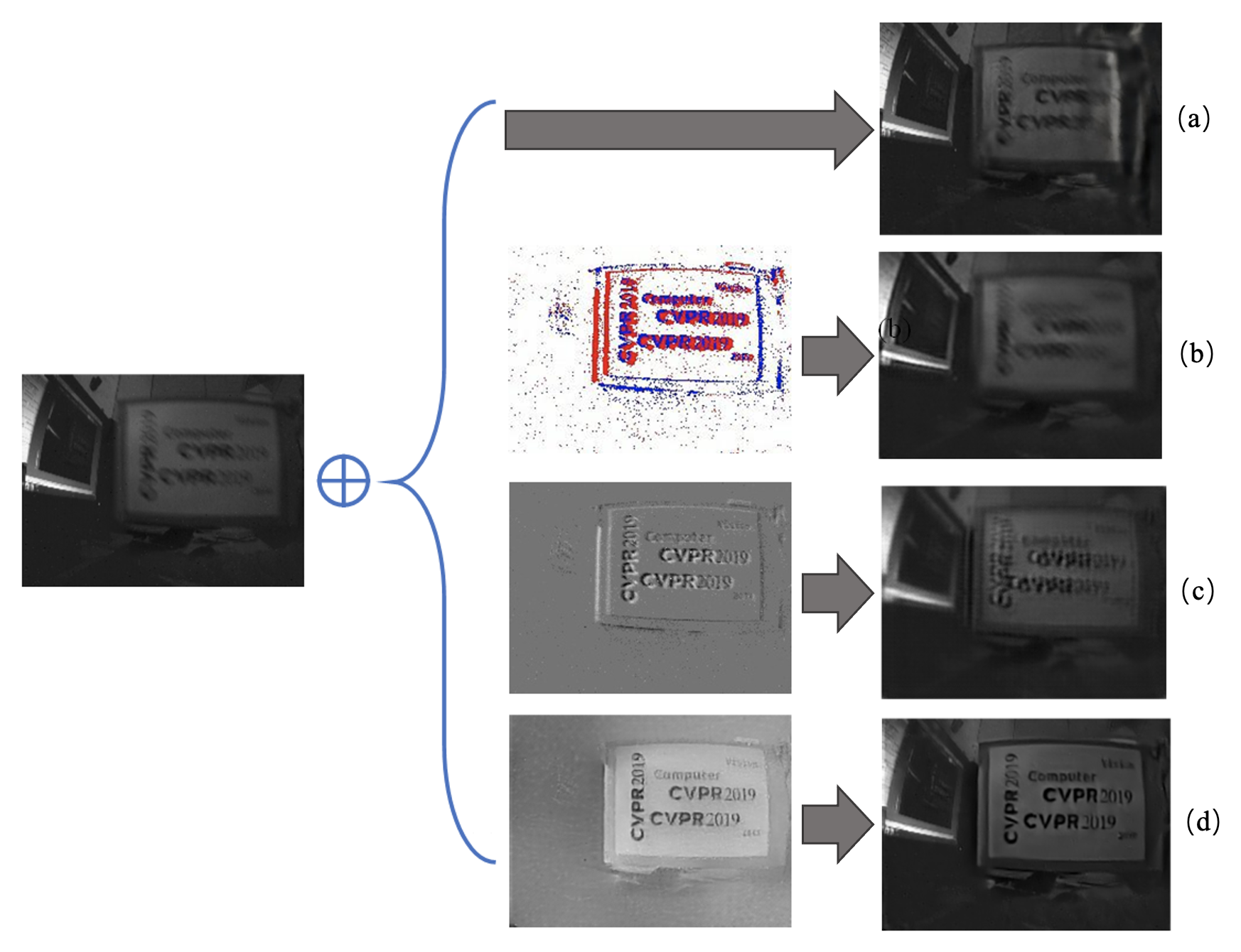}
    \caption{\hg{Various strategies of injecting events into our MPN: (a) directly deblurring of original blurred inputs, (b) deblurring of original blurred input combined with the accumulation of events along the temporal mode (both form the input), (c) deblurring on `original blurred image + event voxel', and (d) deblurring model using the concatenation of original blurred image and Event-to-Video representation.}}
    \label{fig:event_ablation}
\end{figure}

\begin{figure*}[htp]
	\centering
	\includegraphics[width=\linewidth]{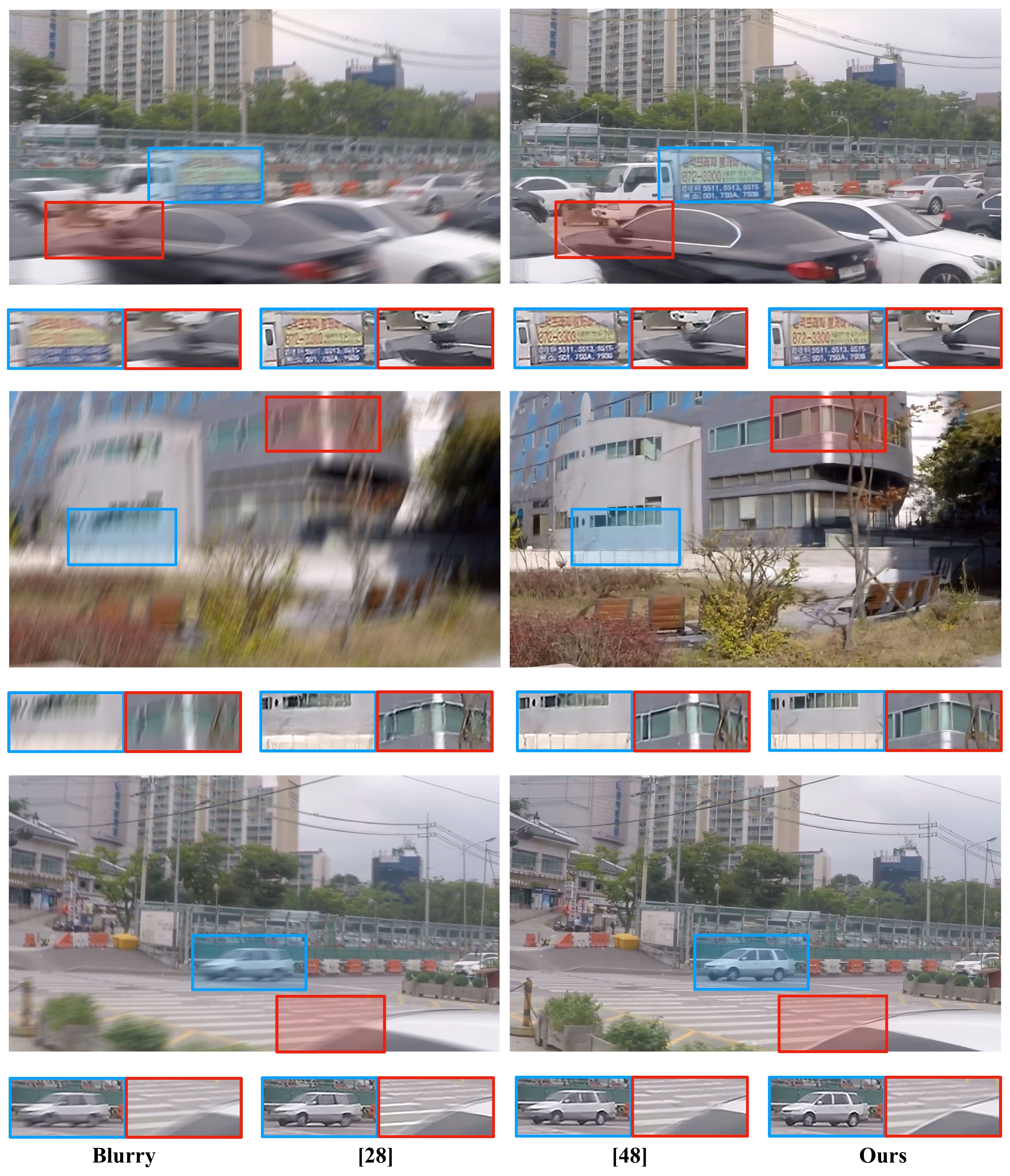}
	\caption{\small Deblurring performance on the blurry images from the GoPro and the VideoDeblurring datasets. We crop regions indicated by blue and red bounding boxes, and obtain smaller windows as follows. The first column contains the original blurry crops, the second column is the result of \cite{nah2017deep}, the third column is the result of \cite{tao2018scale}. Our results are presented in the last column. We argue that our model achieves the best visual performance across several different scenes.}
	\label{fig:Comp2}
\end{figure*}

For video deblurring, using multi-frame inputs does not affect the runtime significantly but it improves the PSNR by 0.41dB. However, as we do not investigate advanced strategies for processing multiple frames, which is out of our focus in this paper, the performance cannot be continuously improved by simply using a larger number of frames. The optimal performance is achieved by using 5 blurry frames.

For event-guided deblurring, we observe that the performance of both image deblurring and video deblurring is significantly boosted. To demonstrate this point, E-MPN achieves $33.14$dB on the GoPro dataset, which outperforms the MPN by up to $\sim$2.9dB. Similar trend is also observed on video deblurring, which is consistent with our theoretical analysis that associating  the event information with blurry images as a composite input to the pipeline should capture accurate motion information, helping the model achieve a better deblurring performance. When we placed the output of the popular TV-L1 optical flow/pretrained FlowNet in place of `event representation' in our pipeline, results of E-StackMPN dropped from 32.57dB to 32.01dB/32.07dB (which is still better than results of \cite{pan2020cascaded,xiang2020deep}) but worse than results of E-StackMPN with `event representation'. While event information may be captured with an extra hardware such as an event camera, the event information used on GoPro dataset in our experiments is entirely simulated from RGB frames by Esim \cite{esim}, which comprises a rendering engine rather than an event camera for ground truth labelling. In case an event camera is available, the quantitative performance of our E-MPN should improve further due to the highest quality of event information in such a case.

Another downside of using the optical flow is that it encodes the displacement information rather than the change information, \etc. Event models can cope with fast motions by design, whereas optical flow algorithms are known to fail under large displacement.

\begin{figure*}[htp]
	\centering
	\includegraphics[width=\linewidth]{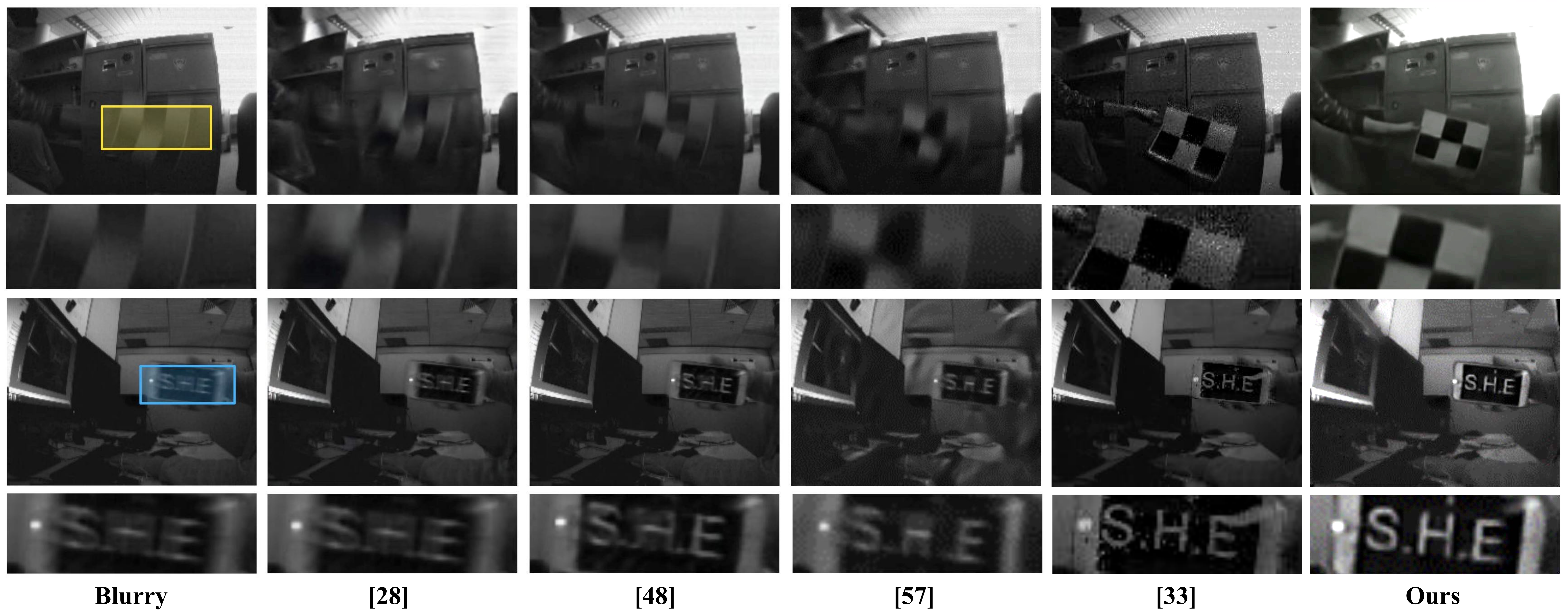}
	\caption{\small \hg{The qualitative comparisons of deblurring performance on blurry images \cite{pan2019bringing}}. The first column contains original blurry images, the second column is obtained with approach \cite{nah2017deep}, the third column is obtained with approach \cite{tao2018scale}, the fourth column is obtained with approach \cite{pan2019bringing}. Our results are presented in the last two columns. The first three deep deblurring models (from left to right) perform poorly when dealing with complex blurs, whereas applying the event representation E-MPN produces crisp images.}
	\label{fig:real}
\end{figure*}

\begin{figure*}[htp]
	\centering
	\includegraphics[width=\linewidth]{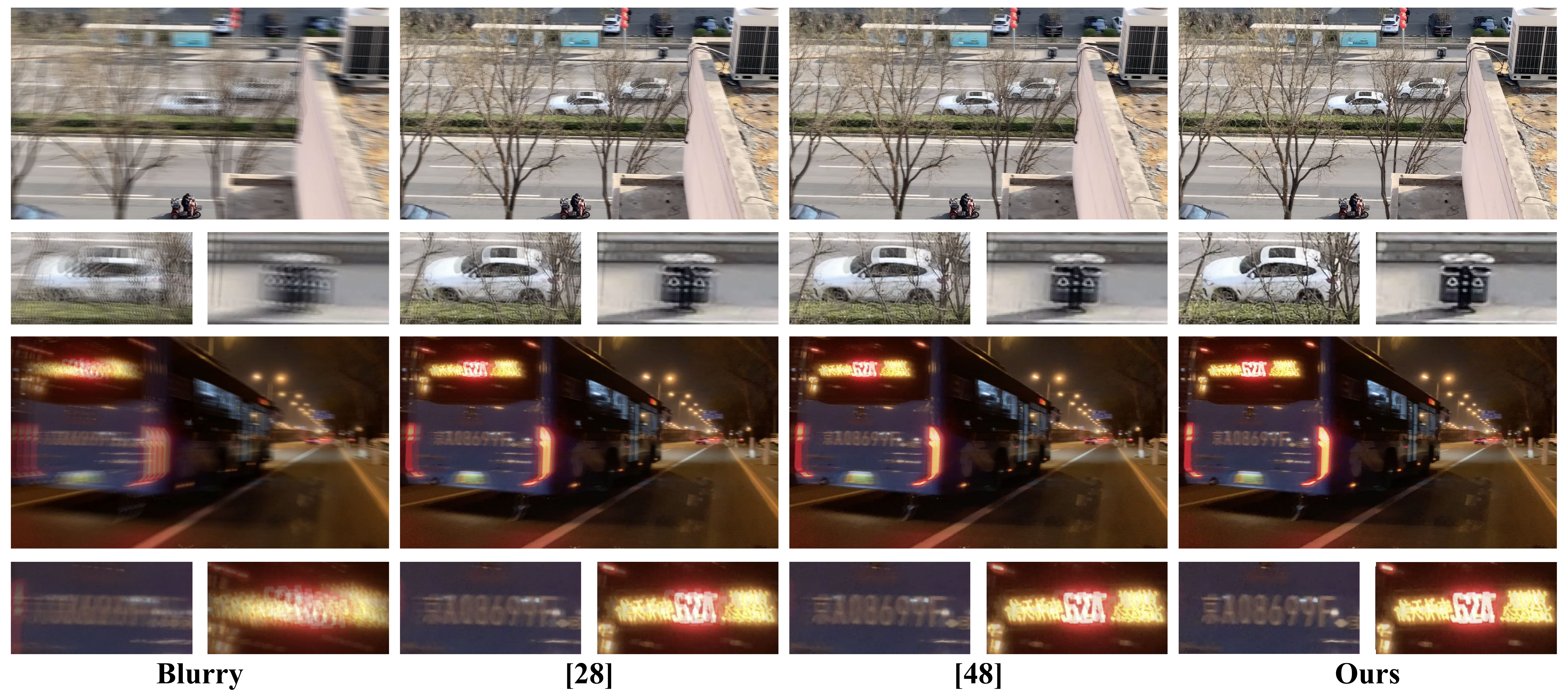}
	\caption{\small \hg{The qualitative comparisons of deblurring performance on realistic blurry images. The first column contains original blurry images. The second column is obtained with approach \cite{nah2017deep}, the third column is obtained with approach \cite{tao2018scale}. Our results are presented in the last column.}}
	\label{fig:real2}
\end{figure*}

The deblurred images from the GoPro dataset are shown in Figures \ref{fig:Comp1}, \ref{fig:Comp2} and \ref{fig:comp_3}. Specifically, Figure \ref{fig:Comp1}  shows the deblurring performance of several models on an image containing heavy a motion blur. We zoom in the main object for clarity. Figure \ref{fig:Comp2} shows selected images of different scenes to demonstrate the advantages of our model which produces the sharpest details across all cases. In addition, we present the deblurring performance on realistic blurry images in Figure \ref{fig:real} to show the benefit of our E-MPN, which clearly outperforms previous deep models in such a scenario. Figure \ref{fig:real2} presents the performance comparison on realistic heavily blurred images consisting of camera and object motions. Our pipeline achieves better deblurring compared to the baseline models.

\noindent\textbf{Runtime}. In addition to the superior PSNR and SSIM of our model, to the best of our knowledge, MPN is also the first deep deblurring model that can work real-time. For example, MPN (1-2-4-8) takes 30ms to process a 720$\times$1280 image, which means it supports real-time 720p image deblurring at 30fps. However, there are runtime overheads related to I/O operations, so real-time deblurring applications require fast transfers from a video grabber to GPU, larger GPU memory and/or an SSD drive, \etc. 

\comment{
\begin{figure}[t]
    \centering
    \includegraphics[width=\linewidth]{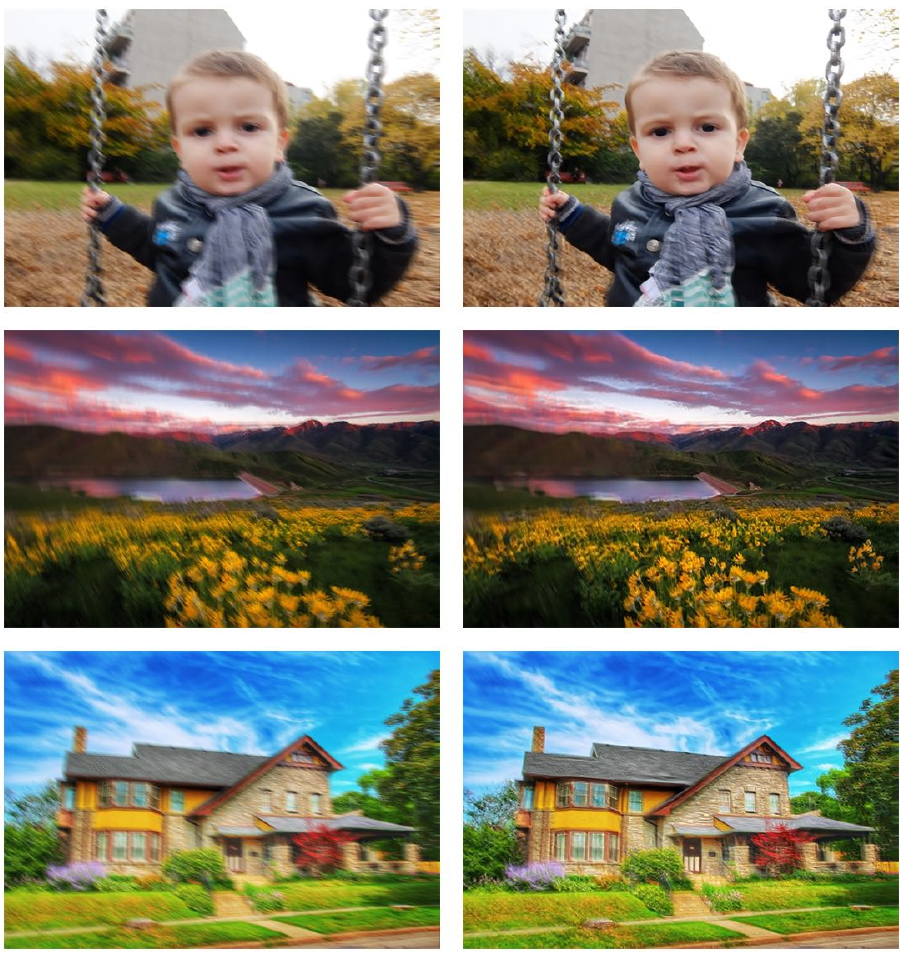}
    \caption{Deblurring instances. The left row shows the original blurry images, whereas the right row shows the sharp deblurred images obtained by SsR-MPN.}
    \label{fig:comp_3}
    \vspace{-0.4cm}
\end{figure}}

The following factors contribute to our fast runtime: i) shallower encoder-decoder with small-size convolutional filters; ii) removal of unnecessary links, \eg, skip or recurrent connections; iii) reduced number of upsampling/deconvolution between convolutional features of different levels. 

\vspace{0.05cm}
\noindent \textbf{Baseline Comparisons.} Despite our model has a much better performance than the multi-scale model \cite{nah2017deep}, it is a somewhat unfair comparison as network architectures of our proposed model and \cite{nah2017deep} differ significantly. Compared with \cite{nah2017deep}, which uses over 303.6MB parameters, we apply much shallower CNN encoders and decoders with the model size 10$\times$ smaller. Thus, we create a deep Multi-Scale Network (MSN) that uses our encoder-decoder following the setup in \cite{nah2017deep} for the baseline comparison (sanity check) between multi-patch and multi-scale methods. As shown in Table \ref{tabel:PSNR_sc}, the PSNR of MSN is worse than \cite{nah2017deep}, which is expected due to our simplified CNN architecture. Compared with our MPN, the best result obtained with MSN is worse than the MPN(1-2) model. Due to the common testbed,  the reported performance of MSN and MPN is the fair comparison of the multi-patch hierarchical and multi-scale models \cite{nah2017deep}.

\subsection{Ablation Studies}
We visualize the outputs of our MPN unit in Figure \ref{fig:dmphn_comp} to analyze intermediate contributions.
As previously alluded to, MPN uses the residual design. Thus, finer levels contain finer but visually less important information compared to the coarser levels. In Fig.~\ref{fig:dmphn_comp}, we illustrate outputs ${\mathbf S}_i$ of each level of MPN (1-2-4-8). The information contained in ${\mathbf S}_4$ is the finest and most sparse. The outputs become less sparse, sharper and richer in color as we move up level-by-level in MPN.

\vspace{0.1cm}
\noindent \textbf{Weight Sharing over Each Level.} Below, we investigate weight sharing between the encoder-decoder pairs of all levels of our network to reduce the number of parameters. Table \ref{tabel:share_weight} shows that weight sharing results in a slight loss of performance but reduces the number of parameters significantly.

\vspace{0.1cm}
\noindent \textbf{The Need for Event Pre-processing.} Our network uses the so-called `events-to-video' simulating network to improve the aggregation of event information and RGB frames. To justify its necessity, we  compare our E-MPN with two recent deep event-guided deblurring models \cite{esl,ecir}, and we find that adding a front-end network to pre-process event voxels is more effective at extracting the motion information from events than other models. Alternatively, a back-end module is required to help compensate for the performance loss but such a module is not easily explainable in the context of using events.

\vspace{0.1cm}
\noindent \textbf{Various Strategies of Injecting Events.}
To justify the necessity of using Event-to-Videos network in our MPN pipeline, we perform ablation studies \wrt different strategies of injecting events. Figure \ref{fig:event_ablation} shows that the last model (concatenation of the original blurred image and Event-to-Video representation) significantly outperforms other baselines in terms of the qualitative comparison, which we attribute to the fact that the Event-to-Videos network decodes events to represent the underlying event/motion dynamics within the image domain rather than the event domain. It is unreasonable to expect a deblurring pipeline could learn decode events by itself, and for that very reason we employ the specialized Event-to-Videos network.

\vspace{0.1cm}
\noindent\textbf{Self-supervised Pipelines.} Combining our proposed self-supervision step  with MPN improves the robustness  \wrt different geometric transformations and photometric  noises, as shown in Table \ref{tabel:robust-comp}. For example, applying a low-level additive Gaussian noise on original blurred images during testing  decreases the deblurring performance of the original model down to 20.98dB, which demonstrates that the original model cannot deblur noisy images. Once the Gaussian noise self-supervision step  is applied,  deblurring performance reaches 30.31dB. Applying rotations as self-supervisory task in MPN  brings around 0.6dB PSNR improvement on GoPro dataset.

\section{Conclusions}
In this paper, we address the challenging problem of non-uniform motion deblurring by exploiting the multi-patch model as opposed to the widely used multi-scale and scale-recurrent architectures. To this end, we have devised an end-to-end deep multi-patch hierarchical deblurring network. Compared against existing deep deblurring frameworks, our model achieves the state-of-the-art performance (according to PSNR and SSIM) and is able to run at 30fps for 720p images. To overcome the discrepancy between adjacent patch boundaries, we explicitly minimize the $\ell_2$ metric between these boundaries to promote the global consistency of patches.  Our stacked variants StackMPN further improve results over both shallower MPN and competing approaches while being $\sim\!\!4\!\times$ faster than the latter models. Our stacking architecture appears to have overcome the limitation to stacking depth which other competing approaches exhibit. Moreover, the novel self-supervised mechanism proposed by us improve the model ability to cope with geometric transformations and photometric noises. Finally, exploiting the camera event representation together with blurred images results in the largest improvements on frames containing complex blur patterns. We hope our work  provides several valuable insights for subsequent works on deblurring.

\begin{sloppypar}
{
\noindent
\textbf{Acknowledgements.}
This research is supported by the Natural Science Foundation of China (Grant No. 62106282). 

\noindent Code: \url{https://github.com/HongguangZhang/DMPHN-cvpr19-master}.

\noindent\textbf{Data availability statement}: All datasets used and studied in this paper are publicly available.
}
\end{sloppypar}

{\small
\bibliographystyle{spmpsci}
\bibliography{Deblur-Reference.bib}
}

\end{document}